\documentclass[letterpaper]{article} % DO NOT CHANGE THIS
\usepackage{aaai24}  % DO NOT CHANGE THIS
\usepackage{times}  % DO NOT CHANGE THIS
\usepackage{helvet}  % DO NOT CHANGE THIS
\usepackage{courier}  % DO NOT CHANGE THIS
\usepackage[hyphens]{url}  % DO NOT CHANGE THIS
\usepackage{graphicx} % DO NOT CHANGE THIS
\usepackage{natbib}  % DO NOT CHANGE THIS AND DO NOT ADD ANY OPTIONS TO IT
\usepackage{caption} % DO NOT CHANGE THIS AND DO NOT ADD ANY OPTIONS TO IT
\usepackage{subfigure}
\usepackage{amsmath}
\usepackage{amsthm}
\usepackage{amssymb}
\usepackage{xcolor}
\usepackage{booktabs}
\usepackage[algo2e,noend]{algorithm2e}
\usepackage{multirow}
\usepackage{siunitx}
\theoremstyle{definition}
\newtheorem{theorem}{Theorem}
\newtheorem{lemma}{Lemma}
\newtheorem{definition}{Definition}
\newcommand{\proofsketch}{\noindent {\bf Proof Sketch: }}
\DeclareMathOperator{\argmax}{arg\,max}
\usepackage{algorithm}
\usepackage{algorithmic}
\usepackage{pdfpages}

\usepackage{newfloat}
\usepackage{listings}
\DeclareCaptionStyle{ruled}{labelfont=normalfont,labelsep=colon,strut=off} % DO NOT CHANGE THIS
\floatstyle{ruled}
\newfloat{listing}{tb}{lst}{}
\floatname{listing}{Listing}
\title{Safe Explicable Planning}
\author{
    Akkamahadevi Hanni,
    Andrew Boateng,
    Yu Zhang
}
\affiliations{
    Arizona State University
    
    ahanni@asu.edu, aoboaten@asu.edu, yzhan442@asu.edu
}

\begin{document}

\maketitle

\begin{abstract}

Human expectations arise from their understanding of others and the world. In the context of human-AI interaction, this understanding may not align with reality, leading to the AI agent failing to meet expectations and compromising team performance. 
Explicable planning, introduced as a method to bridge this gap, aims to reconcile human expectations with the agent's optimal behavior, facilitating interpretable decision-making. However, an unresolved critical issue is ensuring safety in explicable planning, as it could result in explicable behaviors that are unsafe. 
To address this, we propose \textit{Safe Explicable Planning (SEP)}, which extends the prior work to support the specification of a safety bound. 
The goal of SEP is to find behaviors that align with human expectations while adhering to the specified safety criterion. Our approach generalizes the consideration of multiple objectives stemming from multiple models rather than a single model, yielding a Pareto set of safe explicable policies. 
We present both an exact method, guaranteeing finding the Pareto set, and a more efficient greedy method that finds one of the policies in the Pareto set. 
Additionally, we offer approximate solutions based on state aggregation to improve scalability. We provide formal proofs that validate the desired theoretical properties of these methods. 
Evaluation through simulations and physical robot experiments confirms the effectiveness of our approach for safe explicable planning.
\end{abstract}

\section{Introduction}

Significant strides have been made in  advancing the capabilities of AI agents in recent years, from operating in isolated environments to being deployed in environments surrounded by humans. 
Examples of such agents include Starship's food delivery robots, Amazon's Astro household assistants, Bear Robotics' hospitality robots, and Waymo's autonomous vehicles, among many others. 
As technologies evolve, these AI agents are poised to become our indispensable partners. 
It is imperative for AI to learn from human-human interaction where aligning an agent's behavior with others' expectations is a key to such social interaction.

\begin{figure}[t]
    \centering
    \subfigure[Motivating example]
  {\includegraphics[scale=0.18]{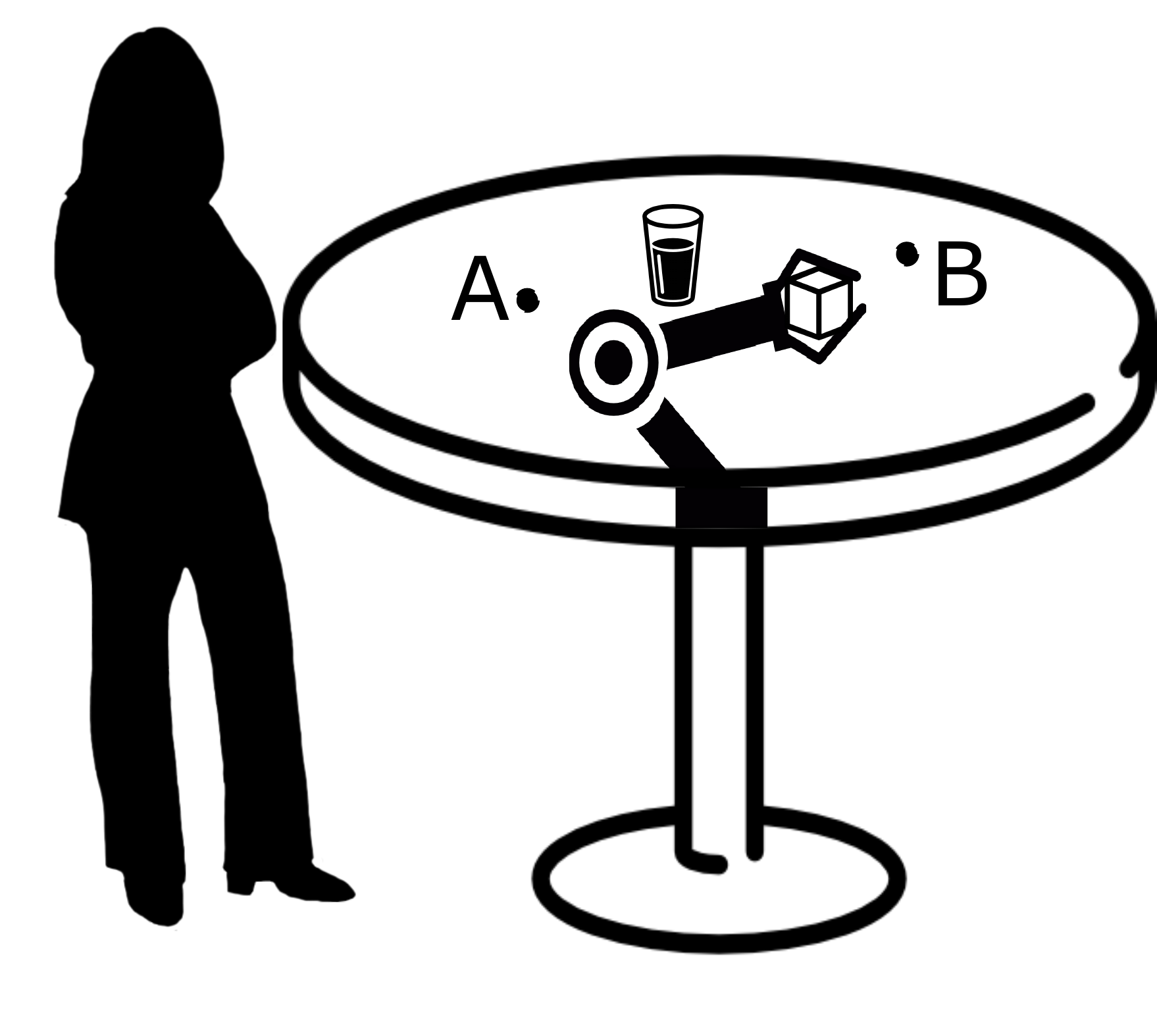}
  \label{fig:motivation}}
    \subfigure[Problem setting]
  {\includegraphics[scale=0.58]{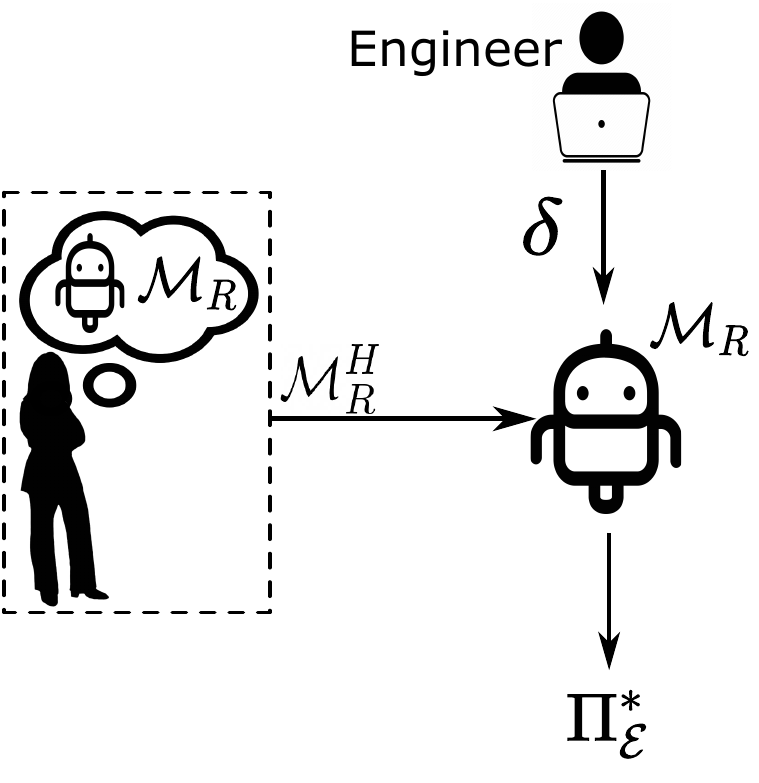}
  \label{fig:SEP-setting}}
    \caption{The agent uses the ground-truth model - $\mathcal{M}_R$, an estimation of the human's understanding of it - $\mathcal{M}_R^H$, and a bound - $\delta$, to generate safe explicable policies $\Pi^*_\mathcal{E}$.}
\end{figure}

Explicable planning \cite{Zhang2017PlanEA,Kulkarni2016ExplicableRP,ActiveExpIROS} is an existing framework addressing human expectations in decision-making. 
It operates under the assumption that humans form their expectations of an agent's behavior based on their perception of the agent and the environment ($\mathcal{M}_R^H$), which may deviate from the reality captured by the agent's model ($\mathcal{M}_R$) (see Fig. \ref{fig:SEP-setting}). 
In the original formulation, the objective is to find a plan that closely resembles the human's expected plan, as measured by an explicability metric, while simultaneously minimizing a plan cost metric through a linearly weighted sum of the two metrics. 
To address explicable behavior generation in stochastic domains, \cite{gong2021explicable} define a similar objective within a learning framework under Markov Decision Processes (MDPs). However, a key drawback of these methods is the lack of consideration of a bound on the suboptimality of the solution under the ground-truth model (i.e., $\mathcal{M}_R$). 
This is due to the fact that the trade-off between cost and explicability metrics (at different scales) is governed by a hyper-parameter, referred to as the reconciliation factor by \cite{Zhang2017PlanEA}. 
Consequently, generating an explicable behavior may overly compromise the cost in the ground-truth model,  leading to potentially unsafe behaviors.\footnote{Here, the underlying assumption is that safety is negatively correlated with the cost metric under $\mathcal{M}_R$. Future work will explore other safety criteria, such as considering behavior variability.}

Let us further illustrate the need for safe explicable planning (SEP) via a motivating scenario. Imagine a human working alongside a robot manipulator. The task is for the robot to hand over a box to the human, with two potential locations for placement: `{\it A}' and `{\it B}' (depicted in Fig. \ref{fig:motivation}). 
Location `{\it A}' is closer to the human but involves a small risk of tipping over a water cup nearby. 
When the cup is empty, this risk is negligible. In such cases, the preferred action would be for the robot to place the box at `{\it A}' to align with the human's expectation. 
However, when the cup is not empty, tipping it over could lead to hazards like electric shocks that incur significant costs in the robot's model. Hence, the preferred action would be for the robot to place the box at `{\it B}'. 
When such a subtle difference (i.e., whether the cup is empty) is not apparent from the human's perspective (based on $\mathcal{M}_R^H$), the robot may indistinguishably prioritize conforming to the human's expectations, leading to unsafe behavior despite seeming more explicable.
In SEP, the robot's behaviors are constrained by a cost bound in the ground-truth model, ensuring it never chooses an unsafe behavior. 
SEP prioritizes safety without sacrificing explicability, which can mitigate the risk by preventing hazardous outcomes in human-robot interaction scenarios.% like the one described above.

In our Safe Explicable Planning (SEP) approach, we build upon the following assumptions to focus on the planning challenges. 
First, we assume that the agent has access to its model ($\mathcal{M}_R$) and the human's belief of its model ($\mathcal{M}_R^H$), or simply, the human's model.
A similar assumption has been made in prior research on explicable planning ~\cite{Kulkarni2016ExplicableRP,ActiveExpIROS} and explainable decision-making ~\cite{chakraborti2019explicability}. 
In practice, the human's model may be provided by experts or acquired from human feedback, which has been explored in previous studies~\cite{christiano2017humanPref, ibarz2018humanPref, holmes2004schema, juba2022learning}.
Second, we assume that the human is a rational observer, i.e., a behavior with a higher expected return in the human's model is more expected.
Hence, the most expected behavior can be generated by computing the optimal behavior in the human's model.
This assumption allows us to equate the problem of maximizing explicability to maximizing the expected return of a policy under the human's model (that is modeled as an MDP).
Such an assumption of human rationality is a common simplification in cognitive science and artificial intelligence research, such as in ~\cite{baker2011bayesian}.

We formulate Safe Explicable Planning (SEP) under MDP by defining the objective as maximizing the expected return in the human's model, subject to a constraint in the agent's model. 
This problem formulation generalizes the consideration of multiple objectives 
% \cite{marler2004survey} 
\cite{white1982multi}
to also consider multiple domain models. The solution to this problem yields a Pareto set of policies for which exact solvers are often intractable.
To address this challenge, we propose an action-pruning technique to reduce the policy space significantly. 
Subsequently, we introduce a novel tree search method that efficiently explores the remaining policies to identify the Pareto set. 
We formally prove that this search method is sound and complete.
Additionally, we introduce a greedy search method for situations where any policy from the Pareto set suffices. 
Finally, we devise approximate solutions for both search methods using state aggregation, addressing scalability in large domains.
We evaluate our methods across several domains via simulation and physical robot experiments, demonstrating their effectiveness for SEP. Furthermore, we conduct ablation studies to analyze the benefits of our pruning techniques, validating their effectiveness in reducing computational costs while generating the desired behaviors.

%%%%%%%%%%%%%%%%%%%%%%%%%%%%%%%%%%%%%%%%%%%%%%%%%%%%%%%%%%%%%%%%%%%%%%%%
\section{RELATED WORK}
%%%%%%%%%%%%%%%%%%%%%%%%%%%%%%%%%%%%%%%%%%%%%%%%%%%%%%%%%%%%%%%%%%%%%%%%

Interest in explainable decision-making has 
been growing with the aim of creating AI agents whose 
behaviors are understandable to humans~\cite{chakraborti2019explicability, chakraborti2020emerging, fox2017explainable}.
We may broadly classify methods in this area into two categories: those that generate interpretable behaviors   (implicit methods) 
and those that communicate to explain behaviors
(explicit methods).
Our work belongs to the former category. 
Researchers have approached implicit methods for explainable decision-making from various but related perspectives, such as generating behaviors that are considered legible \cite{dragan2013generating}, predictable~\cite{dragan2013LegPred}, transparent~\cite{macnally2018action}, explicable~\cite{Zhang2017PlanEA}, etc.
The relationships among these concepts are reviewed comprehensively by~\cite{chakraborti2019explicability}.
Our work extends explicable planning by addressing a 
critical gap in applying such methods to real-world scenarios.

Our problem formulation of SEP shares some key features 
with the constrained-criterion-based formulation of
safe reinforcement learning (RL)~\cite{safeRLsurvey},
which is inherently a Constrained Markov Decision Process (CMDP)~\cite{altman2021CMDP}. Similar problem formulations have been proposed for continuous spaces and applied to risk-bounded motion planning~\cite{huang2019online}.
In these prior works, safety is encoded by constraining the expected cost under some designated cost function while maximizing
the agent's reward function under the same model.
In SEP, similarly, safety is encoded by constraining the expected return under the agent's reward function. 
SEP operates under the assumption that safety directly correlates with the expected return in the agent’s model, following
the intuition that unsafe behaviors would result in low returns.
Our formulation can readily accommodate a CMDP (with a single constraint) 
by aligning the two different models (except for the reward functions) and substituting the robot's reward function in the safety constraint with the cost function.

A distinctive challenge in formulating SEP under CMDP arises from the presence of two different MDP models. Specifically, besides featuring 
two different reward functions, we must explore a more general
setting in SEP that also features two different domain dynamics and discount factors.
This additional complexity makes the existing solution methods for CMDP inapplicable to SEP.
Take, for instance, the linear programming (LP) based approach for CMDP \cite{altman1994denumerable}. 
This method defines the LP objective using an occupation measure for different state-action pairs, which is a function of the transition model and the discount factor. 
However, when dealing with the two different models in SEP, applying the LP solution introduces discrepancies between the occupation measure utilized in the objective and that employed in the constraint. Consequently, resolving these two sets of variables is nontrivial. Similar arguments can be made about the other solution methods.

The objective considered in SEP also bears a similarity to that in Multi-Objective Markov Decision Processes (MOMDP) \cite{white1982multi}, as SEP must consider the expected return under both the agent's and human's model. 
MOMDPs, introduced for multiple objectives under the same MDP (refer to the review paper by \cite{roijers2013survey}), typically aim to optimize a vector of expected returns for those objectives to derive a Pareto set of policies or to derive a single policy through linear scalarization of those objectives. 
Approaches, including but not limited to ~\cite{wakuta1998solution, russell2003q, barrett2008learning, van2013scalarized}, are examples of these methods.
However, to handle different models, MOMDP methods must produce multiple vectors of expected returns, each derived for a different model due to the difference in domain dynamics.
% However, to handle different models, MOMDP solution methods must produce multiple vectors of expected returns, each derived for a different model due to its different domain dynamics.
Optimizing these vectors simultaneously poses a significantly greater challenge than optimizing a single vector in traditional MOMDPs.  

In lexicographic ordered MOMDPs \cite{gabor1998multi, Wray15, pineda2015}, one objective is optimized before the other in a predefined order.
\cite{Wray15} bears close connections to our work and has influenced the action pruning technique outlined in our paper. However, despite the merits, these methods often focus on computational efficiency and do not guarantee the solution's optimality.
In addition, it is unclear how to extend them to handle objectives under different models.

Previous studies have explored solving multiple MDPs \cite{singh1997dynamically, buchholz2019computation}, focusing on identifying a policy that maximizes a combined or weighted sum of objectives,
thus reducing it to a single objective optimization problem.
While these methods may appear comparable to ours, they can yield policies that breach safety bounds or exhibit poor quality in the human's model. 
This drawback stems from their inability to explicitly account for safety constraints, a gap that we address in our work.

%%%%%%%%%%%%%%%%%%%%%%%%%%%%%%%%%%%%%%%%%%%%%%%%%%%%%%%%%%%%%%%%%%%%%%%%
\section{PROBLEM FORMULATION}
%%%%%%%%%%%%%%%%%%%%%%%%%%%%%%%%%%%%%%%%%%%%%%%%%%%%%%%%%%%%%%%%%%%%%%%%

In safe explicable planning, there are two models at play: $\mathcal{M}_R$ and $\mathcal{M}_R^H$. We formulate these models as discrete Markov Decision Processes (MDPs). An MDP is represented by a tuple $\mathcal{M} = \langle \mathcal{S}, \mathcal{A}, \mathcal{T}, \mathcal{R}, \gamma \rangle$ where $\mathcal{S}$ is a set of states, $\mathcal{A}$ is a set of actions, $\mathcal{T}(s' | s, a)$ is a transition function, $\mathcal{R}$ is a reward function,  and $\gamma$ is a discount factor. 
We assume $\mathcal{M}_R$ and $\mathcal{M}_R^H$ share the same state space $\mathcal{S}$ and action space $\mathcal{A}$, but differ in other parameters. Specifically, $\mathcal{M}_R$ incorporates the true domain dynamics $\mathcal{T}_R$, the engineered reward function $\mathcal{R}_R$, and the engineered discount factor $\gamma_R$ whereas $\mathcal{M}_R^H$ incorporates the human's belief about the domain dynamics $\mathcal{T}_R^H$, human's belief about the reward function $\mathcal{R}_R^H$, and human's belief about the discount factor $\gamma_R^H$.
This is reasonable when humans and AI agents coexist in a shared workspace and possess certain shared understanding of the environment.
Relaxing such an assumption incurs separate technical challenges (e.g.,  hierarchical models) that will be deferred to future work. 

We work with the set of all stationary deterministic policies $\Pi$, where $\forall \pi \in \Pi$, $\pi: \mathcal{S} \mapsto \mathcal{A}$.
An agent's optimal policy maximizes the expected return in the agent's model and is given by $\pi^* = \argmax_{\pi} \mathbb{E}_{\mathcal{T}_R}^{\pi} \left[ \sum_{t=0}^{\infty} \gamma_R^t r_R(t) \right]$. 
We define a safe behavior as any behavior with a return within a bound of the agent's optimal return.
Similar criteria have been used in safe RL \cite{safeRLsurvey, moldovan2012deltasafe}. 
More formally, 
% We refer to the designer-specified bound as $\delta \in \mathbb{R}$. The {\textbf{safety criterion}} in our work is defined accordingly: 
a policy $\pi$ is considered safe or feasible if its return satisfies the following condition: %$\forall s^0 \in \mathcal{S}$,
\begin{equation}
    \mathbb{E}_{\mathcal{T}_R}^{\pi} \left[ \sum_{t=0}^{\infty} \gamma_R^t r_R(t) \right] \geq  \delta \mathbb{E}_{\mathcal{T}_R}^{\pi^*} \left[ \sum_{t=0}^{\infty} \gamma_R^t r_R(t) \right], 
    \label{eq:safe}
\end{equation}  
%where $\delta \in \mathbb{R}$ 
where $\delta \in (0, 1]$ 
is the designer-specified safety bound.
Since execution may start from any state, we require such a condition to hold true under {\it any state}. 
It also implies that the condition would hold from any step during execution.
These are desirable features of safety critical systems.

In prior work on explicable planning, the objective is to maximize a weighted sum of the return in the agent's model and an explicability metric. Explicability metric has been defined, for example, via plan distances \cite{Kulkarni2016ExplicableRP} in deterministic domains and KL divergence between trajectory distributions \cite{gong2021explicable} in stochastic domains.
In our work, we define the explicability metric simply as the return in the human's model $\mathcal{M}_R^H$.
Given that the human user generates expectations from $\mathcal{M}_R^H$, this assumes a rational human observer: the higher the return in the human's model, the more expected the policy is. 
\begin{definition}
Safe Explicable Planning (SEP), given by $ \mathcal{P}_\mathcal{E} = \langle \mathcal{M}_R, \mathcal{M}_R^H, \delta \rangle$, is the problem to search for a policy that maximizes the return in $\mathcal{M}_R^H$ subject to a constraint on the return in $\mathcal{M}_R$ under {\it any state}, or formally: 
%$\forall s^0 \in \mathcal{S}$,
\begin{multline}
    \pi^*_\mathcal{E} = \argmax_{\pi} \mathbb{E}_{\mathcal{T}_R^H}^{\pi} \left[ \sum_{t=0}^{\infty} {\gamma^{H}_R}^t {r_R^H}(t) \right] \text{subject to}  \\ 
    \mathbb{E}_{\mathcal{T}_R}^{\pi} \left[ \sum_{t=0}^{\infty} \gamma_R^t r_R(t) \right] \geq \delta \mathbb{E}_{\mathcal{T}_R}^{\pi^*} \left[ \sum_{t=0}^{\infty} \gamma_R^t r_R(t) \right].
    \label{eq:exp_mdp}
\end{multline}
\end{definition}
The maximization of the expected return above across all states introduces a Pareto set of optimal policies where no policies in this set are strictly dominated by any feasible policy.
% Requiring the constraint and optimality above to hold under any state introduces a Pareto set of optimal policies where no policies in this set are strictly dominated by any policy.  
Briefly, a policy $\pi_1$ strictly dominates another policy $\pi_2$ if its state values are no smaller in any state, and larger in at least one state. Formally, we denote such a relationship as $\pi_1 \succ \pi_2$, which holds if $\forall s \in \mathcal{S}$ $[V_{\mathcal{M}_R^H}^{\pi_1} (s) \geq V_{\mathcal{M}_R^H}^{\pi_2} (s)]$ $\wedge$ $\exists s' \in \mathcal{S}$ $[V_{\mathcal{M}_R^H}^{\pi_1} (s') > V_{\mathcal{M}_R^H}^{\pi_2} (s')]$.
The Pareto set $\Pi^*_\mathcal{E}$ is then given by: 
\begin{equation}
     \Pi^*_\mathcal{E} = \{ \pi^*_\mathcal{E} \in \Pi_\delta \hspace{5pt} | \hspace{5pt} \neg \exists \pi \in \Pi_\delta [\pi \succ \pi^*_\mathcal{E}] \},
     \label{eqn:Pi*E}
\end{equation}
where $\Pi_\delta = \{ \pi {\in} \Pi \hspace{1pt} | \hspace{1pt} \forall s {\in} \mathcal{S} \hspace{3pt} [V_{\mathcal{M}_R}^{\pi} (s)  {\geq}  \delta V_{\mathcal{M}_R}^{\pi^*} (s)] \}$ is the set of policies that satisfy the safety bound. 

%%%%%%%%%%%%%%%%%%%%%%%%%%%%%%%%%%%%%%%%%%%%%%%%%%%%%%%%%%%%%%%%%%%%%%%%
\section{SAFE EXPLICABLE PLANNING}
%%%%%%%%%%%%%%%%%%%%%%%%%%%%%%%%%%%%%%%%%%%%%%%%%%%%%%%%%%%%%%%%%%%%%%%%

In this section, we motivate and discuss our solution methods for SEP. 
Given the large policy space to search for, we first discuss a technique to reduce the policy space.  
Since any policy 
$\Pi_{\delta}$ 
may be in the Pareto set, it necessitates the expansion of all policies in $\Pi_{\delta}$
.
We propose an exact method that selectively expands policies in $\Pi_{\delta}$ to determine the Pareto set 
$\Pi^*_\mathcal{E}$
. Additionally, we discuss a greedy method that expands only a subset of policies in 
$\Pi_{\delta}$
, returning a single policy in 
$\Pi^*_\mathcal{E}$
.
Finally, we propose approximate solutions via state aggregation, using handcrafted features, to condition similar states to choose the same actions to further scalability in large domains. Complete proofs, evaluation details, and additional discussions are presented in the full version of this paper~\cite{hanni2024safe}.
%\url{https://arxiv.org/abs/2304.03773}.

%%%%%%%%%%%%%%%%%%%%%%%%%%%%%%%%%%%%%%%%%%%%%%%%%%%%%%
\subsection{Policy Space Reduction via Action Pruning}  
Even though the set $\Pi_\delta$ cannot be obtained directly from the entire policy space $\Pi$, 
we aim to reduce the policy space based on the safety constraint to produce a subset of policies in $\Pi$, referred to as $\widetilde{\Pi}$. 
The challenge here is to ensure that $\widetilde{\Pi} \supseteq \Pi_\delta$  (see Fig. \ref{fig:policySets}).

We achieve this by pruning sub-optimal actions for every state that are guaranteed to violate the constraint. 
Specifically, let $\mathcal{A}(s)$ be the set of all actions that are available in any state $s$. 
The set of actions after pruning is given by: 
\begin{equation}
    \widetilde{\mathcal{A}}(s) {=} \{ a \in \mathcal{A}(s) | 
    Q^{\pi^*}_{\mathcal{M}_R} (s, a) {\geq} \delta \max_{a' \in \mathcal{A}(s)} Q^{\pi^*}_{\mathcal{M}_R} (s, a') \}.
    \label{eq:Abar}
\end{equation}
The policy space obtained from the resulting actions in all states is $\widetilde{\Pi}$. 
Our action pruning technique draws inspiration from \cite{Wray15}. 
In their work, to provide a worst-case guarantee under $\mathcal{M}_R$, the authors employ $1 - (1-\gamma)(1-\delta)$ instead of $\delta$ in Eqn. \eqref{eq:Abar}, resulting in a different set of policies, denoted by $\Pi_\eta$.
Their pruning condition is more stringent than ours and may result in pruning actions prescribed by certain policies that satisfy the constraint in Eqn \eqref{eq:exp_mdp}. 
Consequently, the guarantee that $\Pi_\eta \supseteq \Pi_\delta$ is lost there (see Fig. \ref{fig:policySets}).

\begin{lemma}
    The set of policies after pruning actions based on Eqn. \eqref{eq:Abar} is a superset of the set of policies that satisfy the constraint in Eqn. \eqref{eq:exp_mdp}, i.e., $\widetilde{\Pi} \supseteq \Pi_{\delta}$.
    \label{lemma:l1}
\end{lemma}

\proofsketch
To prove this result, we show that an action pruned in a state per Eqn. \eqref{eq:Abar} is guaranteed to introduce policies that violate the constraint in Eqn. \eqref{eq:exp_mdp} in at least one state. Then, we show the expected return of choosing a pruned action once (in the state it was pruned) and following the optimal policy thereafter, violates the constraint. 
Hence, any policy that chooses the pruned action for that state will result in violating the constraint. 

%%%%%%%%%%%%%%%%%%%%%%%%%%%%%%%%%%%%%%%%%%%%%%%%%%%%%%
\subsection{Policy Descent Tree Search (PDT)} 

To determine $\Pi^*_\mathcal{E}$, 
intuitively, we can evaluate every policy in $\widetilde{\Pi}$.
However, this would be impractical and proves to be unnecessary.  
A more efficient strategy involves further reducing $\widetilde{\Pi}$ by expanding policies in a specific order.
There are two possible search strategies to explore.
First, consider initializing the search to the optimal policy in the human's model and perform policy improvement under the agent's objective until the bound is satisfied. 
Alternatively, consider initializing the search to the optimal policy in the agent's model and perform policy decrement under the agent's objective while simultaneously identifying better policies under the human's objective, until the bound is violated. 
While the first search strategy is simpler it can lead to missed policies in $\Pi^*_\mathcal{E}$, hence we choose the latter option in our work.

In tree search, we start from an optimal policy in $\mathcal{M}_R$, denoted by $\pi^*$, as the root node. 
The benefit of doing so is that, first, we already know that $\pi^*$ satisfies the bound under the agent's model as it is the optimal policy in $\mathcal{M}_R$. 
Second, we can leverage the known state values $V^{\pi^*}_{\mathcal{M}_R}$ to expand policies that have lower state values than that of the parent node, recursively.
Since this is the opposite of policy improvement, we refer to it as policy descent.
Formally, all descendants of a policy $\pi$ under single-action policy updates in PDT can be obtained by replacing $\pi(s)$ under {\it any} state $s$ with an action $a$ that satisfies: 
% \begin{equation}
%      \sum_{s'} \mathcal{T}_R^{}(s, a, s') V_{\mathcal{M}_R}^{\pi} (s') \leq \sum_{s'} \mathcal{T}_R^{}(s, \pi(s), s') V_{\mathcal{M}_R}^{\pi} (s'). 
%      \label{eqn:PDT-condition}
% \end{equation}
\begin{equation}
     Q_{\mathcal{M}_R}^{\pi} (s, a) \leq Q_{\mathcal{M}_R}^{\pi} (s, \pi(s)).
     \label{eqn:PDT-condition}
\end{equation}

\begin{figure}[t]
    \centering
    \subfigure[Policy sets]
    {\includegraphics[scale=0.45]{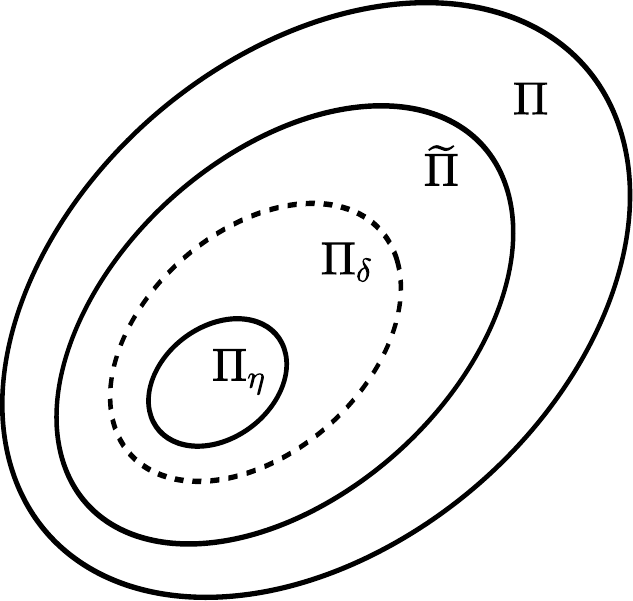}
    \label{fig:policySets}}
    \subfigure[PDT vs. PAG]
    {\includegraphics[scale=0.39]{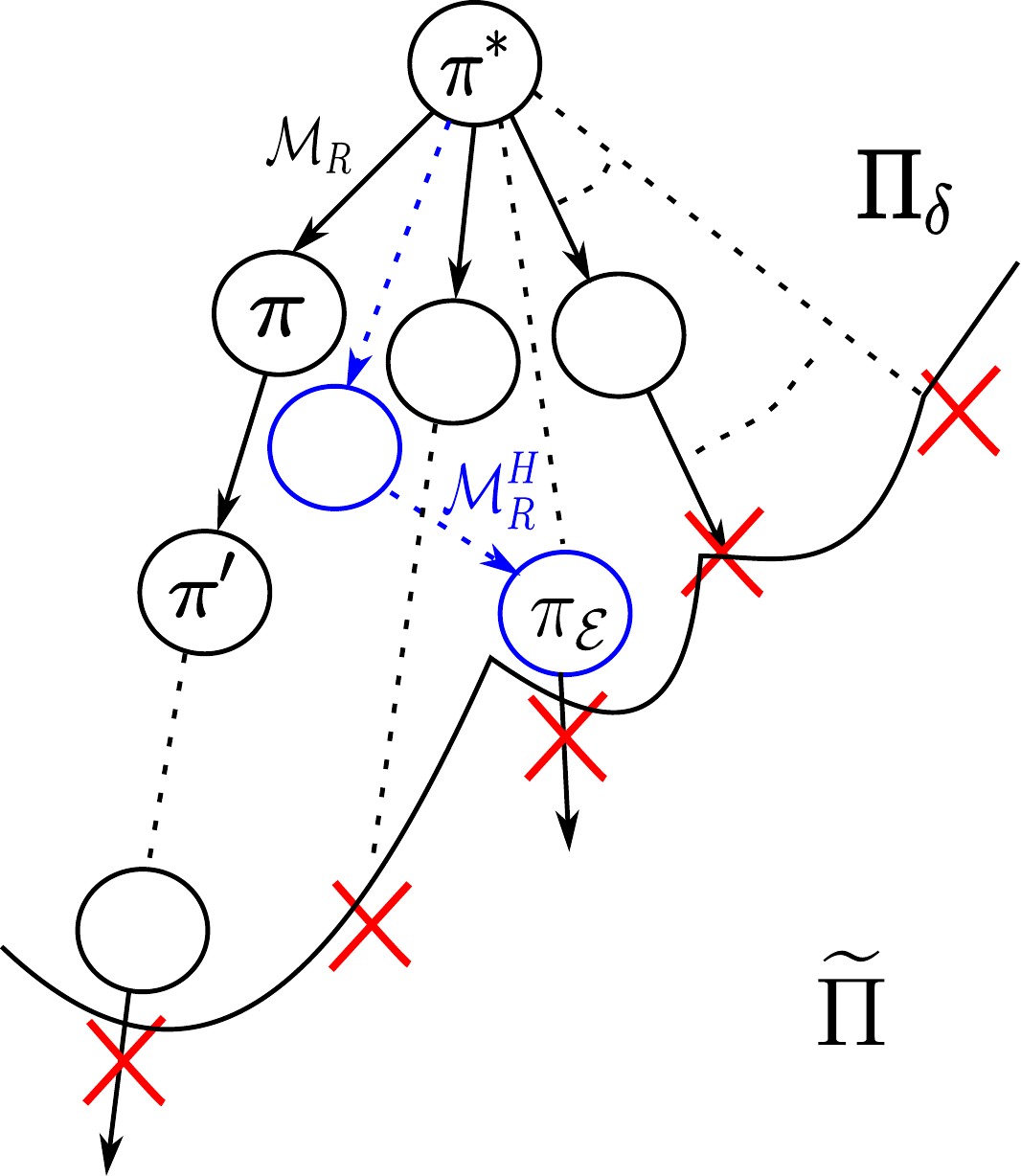}
    \label{fig:tree}}
    \caption{(a) Relationship between different set of policies. 
    (b) PDT vs. PAG on pruned-action space $\widetilde{\Pi}$. The black nodes are expanded by PDT in descending order of state values under $\mathcal{M}_R$. The blue nodes are expanded by PAG in ascending order under $\mathcal{M}_R^H$. Solid lines represent single-action policy updates and dashed links represent multi-action updates.}
\end{figure}

Once a branch reaches a policy whose state values no longer satisfy the bound in $\mathcal{M}_R$ (any state suffices), it is pruned as illustrated in Fig. \ref{fig:tree}. 
The search continues until all branches are explored or pruned while the set of non-dominated policies in $\mathcal{M}_R^H$ are maintained. 
The algorithm is presented in Alg. \ref{alg:PDT+},
which we refer to as PDT+ (includes action pruning).  
Next, we formally show that such a process returns the Pareto set $\Pi_\mathcal{E} = \Pi^*_\mathcal{E}$.

\begin{lemma}
    Let $\pi$ and $\pi'$ be two 
    deterministic 
    policies that differ only by a single action in some state i.e., 
    $\exists s_i \in \mathcal{S}$ [$\pi'(s_i) \neq \pi(s_i)$] $\land$ $\forall s_j \in \mathcal{S} \setminus \{s_i\}$ $[\pi'(s_j) = \pi(s_j)]$ and satisfy 
    $Q^{\pi}_{\mathcal{M}_R}(s_i, \pi'(s_i)) \leq V^{\pi}_{\mathcal{M}_R}(s_i)$. Then, policy $\pi'$ is a descendant of $\pi$ in PDT, i.e., policy $\pi'$ is no better than $\pi$, or formally, 
    $\forall s \in \mathcal{S}$ $[V^{\pi'}_{\mathcal{M}_R}(s) \leq V^{\pi}_{\mathcal{M}_R}(s)]$.
    \label{lemma:l2}
\end{lemma}

\proofsketch
This is an extension of the policy improvement theorem~\cite{sutton2018reinforcement} but in the opposite direction (hence referred to as a policy descent step). 
First, we introduce a temporary non-stationary policy $\pi'_1$ that chooses an action as per $\pi'$ under the initial state and follows $\pi$ thereafter. We can show that the return of $\pi'_1$ is no better than that of $\pi$.  
We can repeat such a pattern to update $\pi'_1$ for the next state and so on, 
resulting in $\pi'$ at the end. 

Similarly, we can show that a special case of the policy improvement theorem holds when a single action is updated (referred to as a policy ascent step). 
% inverse that $\pi$ is an ascendant of $\pi'$ i.e., they satisfy $Q^{\pi'}_{\mathcal{M}_R}(s_i, \pi(s_i)) \geq V^{\pi'}_{\mathcal{M}_R}(s_i)$ and consequently $\forall s \in \mathcal{S}$ $V^{\pi}_{\mathcal{M}_R}(s) \geq V^{\pi'}_{\mathcal{M}_R}(s)$.

% \setlength{\textfloatsep}{0pt}
\begin{algorithm}[t]
\caption{PDT+}
\label{alg:PDT+}
\textbf{Input}: $\mathcal{M}_R$, $\mathcal{M}_R^H$, $\delta$ 
    
$V^*_{\mathcal{M}_R} \gets $ $\mathtt{ValueIteration}(\mathcal{M}_R)$; retrieve $\pi^*$   

\text{Compute} $\widetilde{\mathcal{A}}(s),  \forall s \in S$;

\text{Initialize} $\Pi_\mathcal{E} \gets \emptyset$; $\mathtt{fringe}.\mathtt{push}(\pi^*)$; \\
%Initialize $\mathtt{fringe}.\mathtt{push}(\pi^*)$;
\While{ $\mathtt{fringe} \neq \emptyset$}
{
    $\pi \gets \mathtt{fringe}.\mathtt{pop}()$;
    
    \For{ $a$ in $\widetilde{\mathcal{A}}(s), s \in S $}
    { 
        \If {Eqn. \eqref{eqn:PDT-condition} is satisfied}
        {
            
            $\pi' \gets$ $\mathtt{Modify}(\pi, \pi(s)= a)$;
            
            %\CommentSty{/*$\pi'$ is a descendant of $\pi$*/}

            \If{ $\forall s \in S$ $[V_{\mathcal{M}_R}^{\pi'} (s) \geq \delta V_{\mathcal{M}_R}^{\pi^*} (s)]$}{

            {$\mathtt{fringe}.\mathtt{push}(\pi')$};
            
                \If{$\mathtt{nonDominated}(\pi', \Pi_\mathcal{E}, \mathcal{M}_R^H)$}{
                    
                    $\Pi_\mathcal{E}.\mathtt{update}(\pi')$;
                }
            }
        }
    }
}
\text{return} $\Pi_\mathcal{E}$
\end{algorithm}

\begin{theorem}
PDT+ returns all Pareto optimal policies in $\Pi^*_{\mathcal{E}}$.
\end{theorem}

\proofsketch
To prove this, we show that there exists a policy descent path from any optimal policy (denoted by $\pi^*$) in $\mathcal{M}_R$ (i.e., the root node in PDT) to any Pareto optimal policy (denoted by $\pi^*_{\mathcal{E}}$) in $\Pi^*_{\mathcal{E}}$ by induction. When $\pi^*_{\mathcal{E}}$ differs from $\pi^*$ in only $1$ action, $\pi^*_{\mathcal{E}}$ must be one of the direct descendants of $\pi^*$ in PDT as $\pi^*$ is the optimal policy in $\mathcal{M}_R$.
Hence, $\pi^*_{\mathcal{E}}$ will be expanded by PDT. 
Assume any policy that differs from $\pi^*$ in $k$ actions is expanded. 
When $\pi^*_{\mathcal{E}}$ differs from $\pi^*$ in $k+1$ actions, we show that there must exist a policy $\pi$ that differs from $\pi^*$ in $k$ out of the $k+1$ actions (hence differing from $\pi^*_{\mathcal{E}}$ in $1$ action) and (by Lem. \ref{lemma:l2}) is no worse than $\pi^*_{\mathcal{E}}$ under $\mathcal{M}_R$  via proof by contradiction. 
Consequently, $\pi^*_{\mathcal{E}}$ must be a descendant of $\pi$ in PDT. 
Since $\pi$ is expanded under our inductive assumption, $\pi^*_{\mathcal{E}}$ will be expanded.
Then by Lem. \ref{lemma:l1}, the conclusion holds.

%%%%%%%%%%%%%%%%%%%%%%%%%%%%%%%%%%%%%%%%%%%%%%%%%%%%%%
\subsection{Policy Ascent Greedy Search (PAG)}
In certain situations, it may be unnecessary to compute  $\Pi^*_\mathcal{E}$: any policy in the set would suffice. 
To this end, we introduce a greedy method that only searches through a subset of $\Pi_\delta$, 
making it computationally more efficient than PDT. 

Similar to PDT, we start with $\pi^*$ at the root node.
However, unlike in PDT where we expand policies that have lower state values in $\mathcal{M}_R$ via single-action policy updates, we expand only a single policy that has higher values in $\mathcal{M}_R^H$ than its parent node via multi-action policy updates (see Fig. \ref{fig:tree}). 
Formally, only one descendant of policy $\pi$ is expanded in PAG, which is obtained by replacing $\pi(s)$ under {\it each} state $s$ with an action $a$ that satisfies the following condition (similar to a policy improvement step): 
% \begin{equation}
%     \sum_{s'} \mathcal{T}_R^{H}(s, a, s') V_{\mathcal{M}_R^H}^{\pi} (s') \geq \sum_{s'} \mathcal{T}_R^{H}(s, \pi(s), s') V_{\mathcal{M}_R^H}^{\pi} (s'), 
%     \label{eqn:PAG-condition}
% \end{equation}
\begin{equation}
    Q_{\mathcal{M}_R^H}^{\pi} (s, a) \geq Q_{\mathcal{M}_R^H}^{\pi} (s, \pi(s)),
    \label{eqn:PAG-condition}
\end{equation}
where each such state-action update is checked against the constraint in Eqn. \eqref{eq:exp_mdp} (in $\mathcal{M}_R$) incrementally and incorporated only if the constraint is not violated, resulting in a multi-action policy update for $V_{\mathcal{M}_R^H}^{\pi}$. 
% \begin{equation}
%     \sum_{s'} \mathcal{T}_R^{H}(s, a, s') V_{\mathcal{M}_R^H}^{\pi^*_{\mathcal{E}}} (s') \geq \sum_{s'} \mathcal{T}_R^{H}(s, \pi^*_{\mathcal{E}}(s), s') V_{\mathcal{M}_R^H}^{\pi^*_{\mathcal{E}}} (s')
%     \label{eqn:PAG-condition}
% \end{equation}

% When there are multiple state-action updates,
% each update
% where each such state-action update checked against the constraint in Eqn. \eqref{eq:exp_mdp}

In PAG, we maintain a single candidate policy $\pi_\mathcal{E}$ 
as opposed to a set in PDT. % $\Pi^*_\mathcal{E}$. 
The current policy $\pi_\mathcal{E}$ 
is updated to its descendant $\pi'$ if at least one of the state-action updates is incorporated.  
This process is repeated until $\pi_\mathcal{E}$ remains unchanged. The algorithm, referred to as PAG+ (includes action pruning), is presented in Alg. \ref{alg:PAG+}.

\begin{algorithm}[t]
\caption{PAG+}
\label{alg:PAG+}
\textbf{Input}: $\mathcal{M}_R$, $\mathcal{M}_R^H$, $\delta$

$V^*_{\mathcal{M}_R} \gets $ $\mathtt{ValueIteration}(\mathcal{M}_R)$; retrieve $\pi^*$   

\text{Compute} $\widetilde{\mathcal{A}}(s),  \forall s \in S$;

\text{Initialize} $\pi_{\mathcal{E}} \gets \pi^*$; $\mathtt{changed}$ $\gets true$;

\While{ $\mathtt{changed}$ }{
    
    $V_{\mathcal{M}_R^H}^{\pi_{\mathcal{E}}} \gets $ $\mathtt{PolicyEvaluation}(\pi_{\mathcal{E}}, \mathcal{M}_R^H)$
    
    $\mathtt{changed}$ $\gets false$
    
    \For{ $a$ in $\widetilde{\mathcal{A}}(s), s \in S $}{ 
        
        \If{Eqn. \eqref{eqn:PAG-condition} is satisfied}{

            $\pi' \gets$ $\mathtt{Modify}(\pi_{\mathcal{E}}, \pi_{\mathcal{E}}(s)= a)$;
            
            \If{ $\forall s \in S  [V_{\mathcal{M}_R}^{\pi'} (s) \geq \delta V_{\mathcal{M}_R}^{\pi^*} (s)]$}{
                
                Update $\pi_{\mathcal{E}} \gets \pi'$
                
                $\mathtt{changed}$ $\gets true$
            }
        }

    }
}
return $\pi_{\mathcal{E}}$

\end{algorithm}

\begin{theorem}
PAG+ returns a policy in the Pareto set $\Pi^*_\mathcal{E}$.
\end{theorem}

\proofsketch
The PAG search process stops when it can no longer improve or find a policy that is equivalent in values to $\pi_\mathcal{E}$ under $\mathcal{M}_R^H$ while satisfying the safety constraint. 
This implies that there does not exist a state-action update that implements a policy ascent step under the constraint. 
However, if $\pi_\mathcal{E} \notin \Pi^*_\mathcal{E}$, 
there must exist another policy $\pi \in \Pi^*_\mathcal{E}$ that dominates $\pi_\mathcal{E}$,
which contradicts with the fact that no policy ascent step exists. 
Then by Lem. \ref{lemma:l1}, $\pi_\mathcal{E} \in \Pi^*_\mathcal{E}$.

\subsection{Approximate Solution via State Aggregation}

In the worst-case scenario, both PDT+ and PAG+ must explore a number of policies on the order of $|\widetilde{\Pi}|$, which remains exponential. Consequently, directly applying these methods to complex domains proves challenging. Approximate solutions become essential. However, conventional methods relying on function approximation for state value functions to search for optimal policies~\cite{sykora2008state,abel2016nearOptimal,abel2018lifelongRL,ferrer2020kmdps} are not applicable here, as the search is conducted over the policy space.

We aim to devise an approximate solution that minimizes the number of unique policies to be explored. 
Inspired by function approximation, one approach is to condense the state space by grouping together states that exhibit similar action selection tendencies. 
The similarity of states can be measured using domain-specific features. 
By conditioning states within the same clusters to select the same actions under any policy with either model, we effectively reduce the state space size and consequently the number of policies. 
Formally, this process involves introducing a mapping $\Phi: \mathcal{S}_K \mapsto \mathcal{S}$, establishing a one-to-many correspondence from clusters to states, where $K$ denotes the number of clusters. 
Both PDT and PAG can operate using the aggregated state space (i.e., clusters), treating $\mathcal{S}_K$ as the new state space.

Under the assumption that the states within any cluster are ``correlated'' in action selection under any given policy, the theoretical guarantees of optimality, completeness, and constraint satisfaction remain intact.  
Such a situation may occur, for example, when two states are topologically equivalent, such that a reasonable policy should always choose the same action under these states. 
Investigating the introduction of such states and their impact on guarantees when this assumption does not hold or holds only approximately would be interesting. 
From this perspective, our approximation method resembles function approximation in Q-learning.

\section{EVALUATION}

We evaluate our methods across various domains through simulation and physical robot experiments, aiming to achieve three main objectives. 
First, we compare safe explicable behaviors with optimal behaviors to validate the efficacy of our approach.
Second, given that solving SEP involves searching for the optimal policy in the feasible policy space to obtain the Pareto set, we evaluate the efficiency of our proposed methods and compare them with baselines (BF \& BF+) that employ brute-force policy search. 
Notably, our comparisons are against brute-force methods because prior studies discussed in the related work section lack consideration for multiple models or safety bounds (refer to related work). 
Additionally, we conduct ablation studies for each proposed method to analyze the benefits of pruning actions and our approximate solutions in more complex domains.
Third, we conduct physical robot experiments to demonstrate the applicability of our approach to real-world scenarios. 
Following our naming convention, we append '+' to an algorithm's name to denote the incorporation of our action pruning technique, resulting in a reduced policy space $\widetilde{\Pi}$; a method without '+' must search the original policy space, or $\Pi$ (refer to Fig. \ref{fig:policySets}).
All evaluations were run on a MacBook Pro (16 GB, 3.1 GHz Dual-Core Intel Core i5).

{\it Bound Selection}: In our approach, we assume the bound is specified by the designer, based on experience. 
However, it can often be estimated based on the domain. For instance, consider one of the cliff worlds depicted in Fig. \ref{fig:cliff-paretoSet} (see below). 
The optimal return in the agent's model is $94$ (i.e., moving along the edge of the cliff to the goal), while the return of the trajectory with the longest detour (i.e., staying as far away from the edge as possible) without falling off the cliff is $90$, discount notwithstanding. 
As unsafe behaviors yield significantly lower returns (in $\mathcal{M}_R$) than the detour, the safety bound can be set to $90/94 {=} 0.957$, subject to adjustment. Further analyses will be deferred to future work.

{\it Policy Selection}:
 User preferences can serve as a guiding factor to select from the Pareto set $\Pi^*_{\mathcal{E}}$. Alternatively, domain-specific scores may be introduced to aid in the selection process. For instance, policies that are deemed ``simpler'' may receive higher scores. Example scores are discussed below, wherever applicable.

\subsection{Simulations}

\subsubsection{Domain Descriptions:}

$1)$ {\it Cliff Worlds (CS \& CL)}: 
The task entails navigating alongside the edge of a cliff to reach the goal, as depicted in Figs. \ref{fig:cliff} and \ref{fig:cliff-paretoSet}. 
The ground-truth ($\mathcal{M}_R$) is that the agent can travel alongside the edge without slipping off the cliff. 
Conversely, the human's belief ($\mathcal{M}_R^H$) is that there is a probability that the agent may slip off from the edge, especially in terrain closer to the cliff, which is more uneven and challenging to traverse. 
Both models have similarly defined reward functions, $\mathcal{R}_R$ and $\mathcal{R}_R^H$, shown in Figs. \ref{fig:cliff:pi*_MR} and \ref{fig:cliff:pi*_MH}, respectively, for the larger domain.
We designed a small $4 \times 5$ domain (CS) for the exact methods and a large $4 \times 100$ domain (CL) for approximate solutions. 
To apply approximate solutions to CL, we aggregated all non-terminal states based on features such as distance to the cliff and the agent's position in the grid (e.g., along the edge or at the ends) into 10 clusters while retaining the terminal states as they are.

\begin{figure*}[ht]
\centering
\subfigure[$\tau(\pi^*_{\mathcal{M}_R})$]{\includegraphics[scale=0.23]{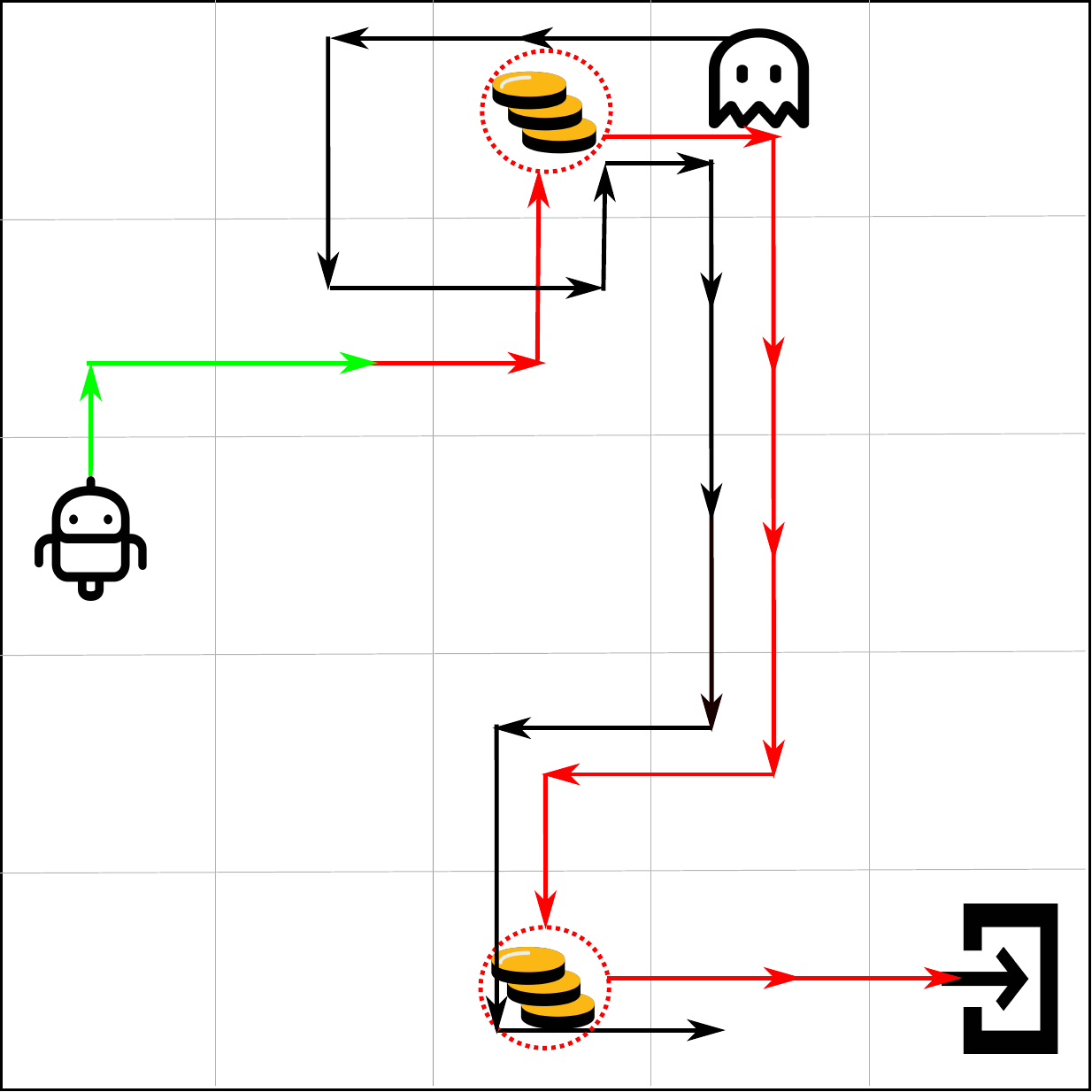}\label{fig:wump:pi*_R}}
\subfigure[$\tau(\pi^*_{\mathcal{M}_R^H})$]{\includegraphics[scale=0.23]{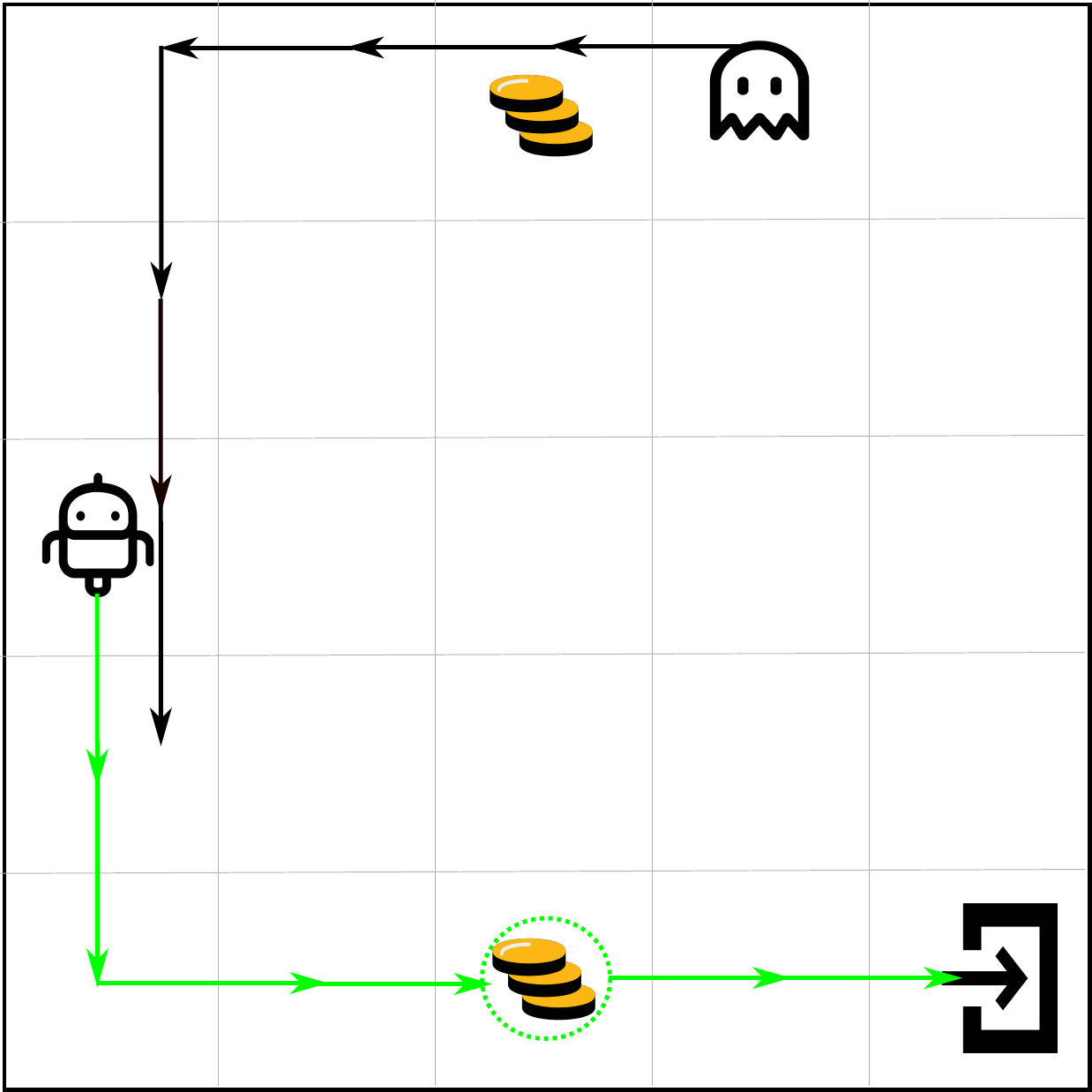}\label{fig:wump:pi*_H}}
\subfigure[$\tau(\pi^*_\mathcal{E})$ $|$ PDT+]{\includegraphics[scale=0.23]{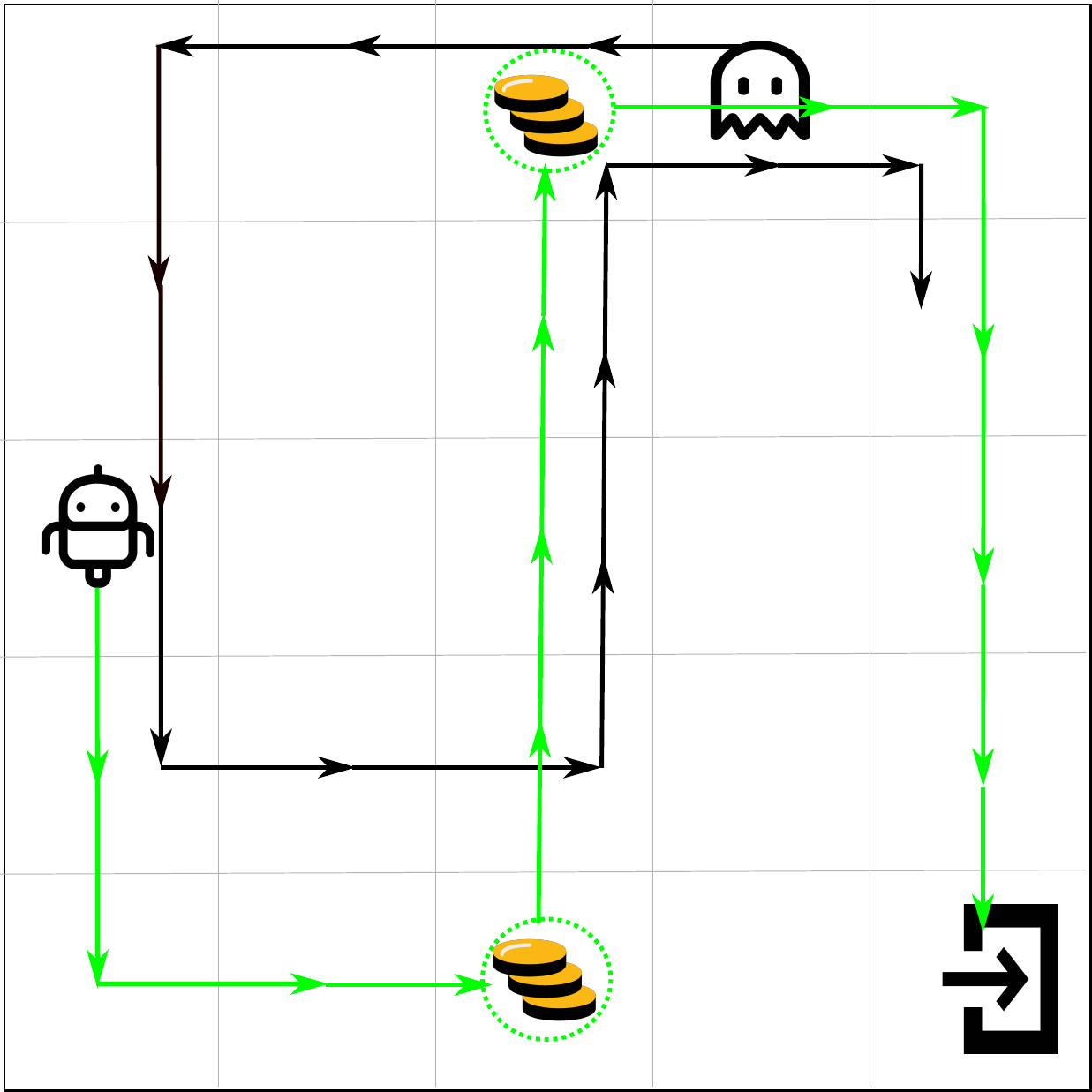}\label{fig:wump-PDT}}
\subfigure[$\tau(\pi^*_\mathcal{E})$ $|$ PAG+]{\includegraphics[scale=0.23]{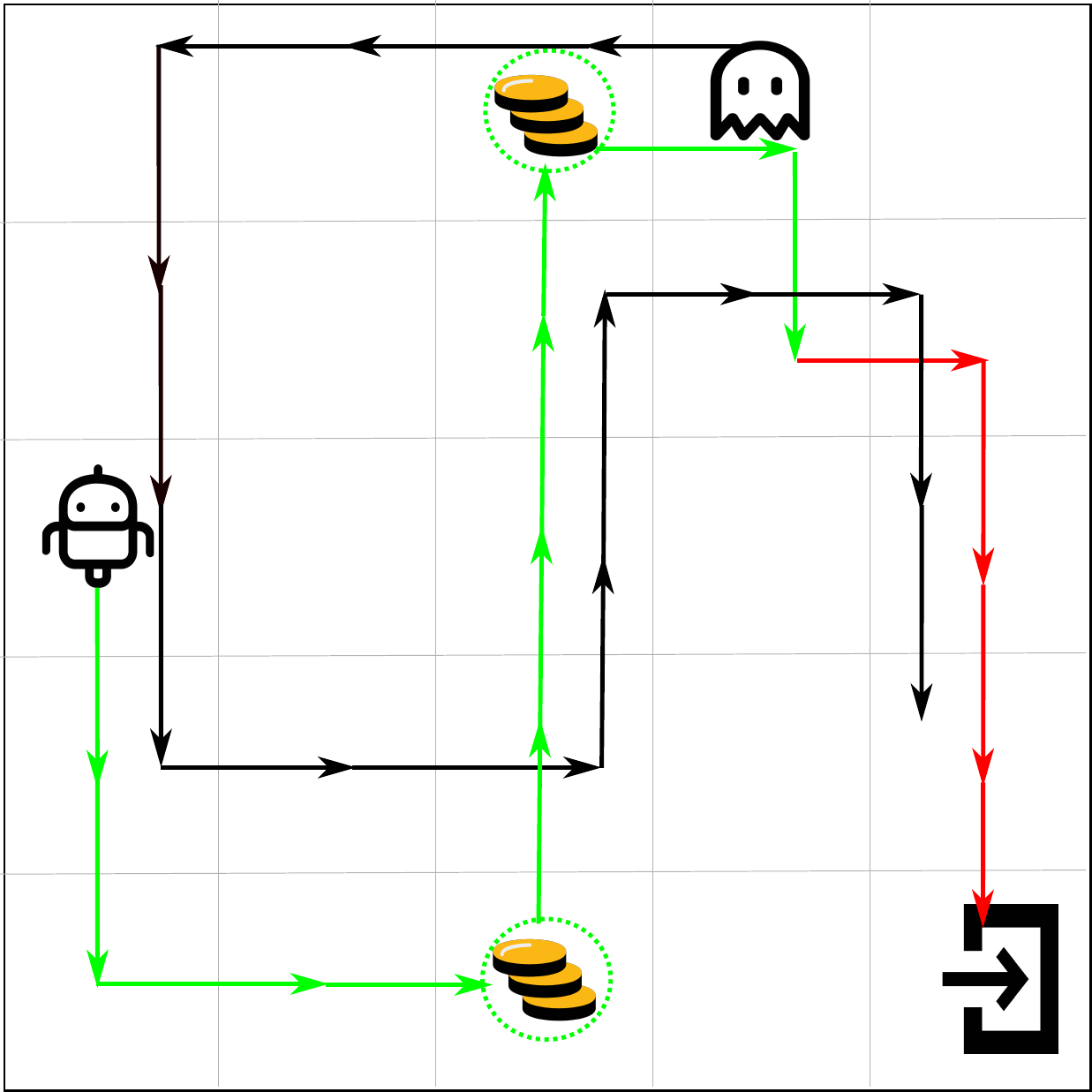}\label{fig:wump-PAG}}
\caption{Behavior comparison in the wumpus world. Black lines show the trajectories of the wumpus. Red line segments show the parts of the agent's trajectories when the wumpus is in an adjacent cell, and green line segments show when the wumpus is at least two steps away. 
Presented are the most likely trajectories by (a) the optimal agent's policy, (b) the human's expectation, and the safe explicable policies obtained when $\delta = 0.90$ by (c) PDT+ and (d) PAG+,  respectively.}
\label{fig:wumpus}
\end{figure*}

$2)$ {\it Wumpus World (W)}:
The agent's objective is to exit a $5 \times 5$ cave while collecting gold coins and avoiding encounters with the wumpus (i.e., entering the same location as the wumpus) (see Fig. \ref{fig:wumpus}). 
The wumpus always moves towards the agent. 
Each collected gold coin yields a reward of $+30$, and the game ends if the agent encounters the wumpus ($-100$) or exits the cave ($+100$). 
The ground-truth ($\mathcal{M}_R$) is that the agent's actions are deterministic, whereas the wumpus's actions are stochastic. 
Conversely, the human's belief ($\mathcal{M}_R^H$) is that both agents' actions are stochastic. 
Under this belief, the human perceives it as risky for the agent to approach the wumpus.
For approximate solutions, non-terminal states were aggregated into 15 clusters based on features such as the relative direction of the wumpus from the agent and collected gold coin(s).

%%%%% WUMPUS EXPERIMENT PICTURES

\subsubsection{Results:}

%%%%%%%
% Ablation Experiment
\begin{table}[!t]
    \centering
    \setlength\tabcolsep{0.4 pt}
    \begin{tabular}{|c|c|c|c|r|r|r|r|r|r|r|r|r|}
        \hline
            & $\delta$ & BF & BF+ & \multicolumn{2}{c|}{PDT} & \multicolumn{2}{c|}{PDT+} & $|\Pi^*_{\mathcal{E}}|$ & \multicolumn{2}{c|}{PAG} & \multicolumn{2}{c|}{PAG+} \\ 
        \cline{3-13} 
            &          & \# & \#      &  \multicolumn{1}{|c|}{\#} & \multicolumn{1}{|c|}{RT}  &  \multicolumn{1}{|c|}{\#}   & \multicolumn{1}{|c|}{RT}  &   & \multicolumn{1}{|c|}{\#} & \multicolumn{1}{|c|}{RT} & \multicolumn{1}{|c|}{\#} & \multicolumn{1}{|c|}{RT} \\
        \hline
                & $1.00$ & $4^{16}$ & $4^{4\ }$       & $8448$        & $4.8$   & $256$         & $0.3$     & $1$  & $17$ & $0.01$ &  $9$ & $0.01$ \\
                & $0.95$ & $4^{16}$ & $\dot{4}^{9\ }$ & $8448$        & $4.8$   & $2816$        & $1.7$     & $1$  & $17$ & $0.01$ & $10$ & $0.01$ \\
        CS      & $0.93$ & $4^{16}$ & $\dot{4}^{15}$  & $8448$        & $4.8$   & $7424$        & $4.3$     & $1$  & $17$ & $0.01$ & $17$ & $0.01$ \\
                & $0.90$ & $4^{16}$ & $\dot{4}^{15}$  & $16\dot{9}$k  & $102.2$ & $14\dot{9}$k  & $90.3$    & $3$  & $21$ & $0.02$ & $19$ & $0.01$ \\
                & $0.85$ & $4^{16}$ & $\dot{4}^{15}$  & $31\dot{3}$k  & $184.4$ & $27\dot{4}$k  & $164.2$   & $3$  & $19$ & $0.01$ & $19$ & $0.01$ \\
        \hline
                & $1.00$ & $4^{10}$ & $4^{2\ }$       & $368$   & $5.0$  & $16$     & $0.5$    & $1$ & $36$ & $0.5$  &  $5$ & $0.1$  \\
                & $0.97$ & $4^{10}$ & $\dot{4}^{9\ }$ & $684$   & $7.9$  & $620$    & $7.1$    & $1$ & $36$ & $0.5$  & $32$ & $0.4$  \\
        CL      & $0.95$ & $4^{10}$ & $\dot{4}^{9\ }$ & $1846$  & $21.0$ & $1677$   & $18.2$   & $3$ & $33$ & $0.5$  & $30$ & $0.4$  \\
                & $0.93$ & $4^{10}$ & $\dot{4}^{9\ }$ & $2254$  & $25.1$ & $2048$   & $22.8$   & $2$ & $30$ & $0.4$  & $27$ & $0.4$  \\
                & $0.90$ & $4^{10}$ & $\dot{4}^{9\ }$ & $2268$  & $25.4$ & $2060$   & $25.0$   & $2$ & $30$ & $0.5$  & $27$ & $0.4$  \\
        \hline
                & $1.00$ & $4^{15}$ & $4^{0\ }$         & $61$          & $0.9$   & $1$     & $0.1$   & $1$   & $25$ & $0.4$  & $1$ & $0.1$  \\
                & $0.97$ & $4^{15}$ & $\dot{4}^{1\ }$   & $61$          & $0.9$   & $3$     & $0.1$   & $1$   & $25$ & $0.4$  & $1$ & $0.1$  \\
        W       & $0.95$ & $4^{15}$ & $\dot{4}^{5\ }$   & $61$          & $0.9$   & $13$    & $0.3$   & $1$   & $25$ & $0.4$  & $5$ & $0.1$  \\
                & $0.93$ & $4^{15}$ & $\dot{4}^{5\ }$   & $1489$        & $21.7$  & $179$   & $4.1$   & $25$  & $46$ & $0.7$  & $5$ & $0.2$  \\
                & $0.90$ & $4^{15}$ & $\dot{4}^{5\ }$   & $2\dot{4}$k   & $359.1$ & $729$   & $42.1$  & $197$ & $46$ & $0.7$  & $5$ & $0.2$  \\
        \hline
    \end{tabular}
    \caption{Comparison of different methods via the number of policies evaluated (\#) and runtime (RT) in minutes. Numbers with a dot are approximate.}
    \label{tab:ablation&runtime}
\end{table}

$1)$ {\it Performance Comparison}:
Table \ref{tab:ablation&runtime} presents the runtime (except for BF and BF+ due to the large number of policies) and the number of policies expanded (or evaluated) by each method across the three simulation domains.
The small cliff world (CS) comprises $16$ non-terminal states and $4$ actions in each state, resulting in $|\Pi| {=} 4^{16}$ policies.
The large cliff world (CL) comprises $301$ non-terminal states and $4$ actions in each state, resulting in $|\Pi| {=} 4^{301}$ policies, but upon aggregation, it reduces to $4^{10}$ policies.
The wumpus world (W) comprises $2116$ non-terminal states and $4$ actions in every state, resulting in $|\Pi| {=} 4^{2116}$ policies, but upon aggregation, it reduces to $4^{15}$ policies.
We employ approximate solutions (i.e., PDTs and PAGs on aggregated state space) on CL and W.

We observe that action pruning effectively reduces the policy space and consequently the number of policies expanded by BF+, PDT+, and PAG+ compared to BF, PDT, and PAG, respectively. 
The expansion order of policies in PDTs leads to significant additional reduction compared to BF+. 
With or without action pruning, PAGs expand fewer policies than PDTs as they only need to return a single policy. 
Lastly, although the number of policies expanded in PDTs increases (for lower $\delta$), it is worth noting that PAGs sometimes expand fewer policies due to their greedy nature.

\begin{figure*}[ht!]
  \centering

  \subfigure[$\tau(\pi^*_{\mathcal{M}_R})$]
  {\includegraphics[scale=0.25]{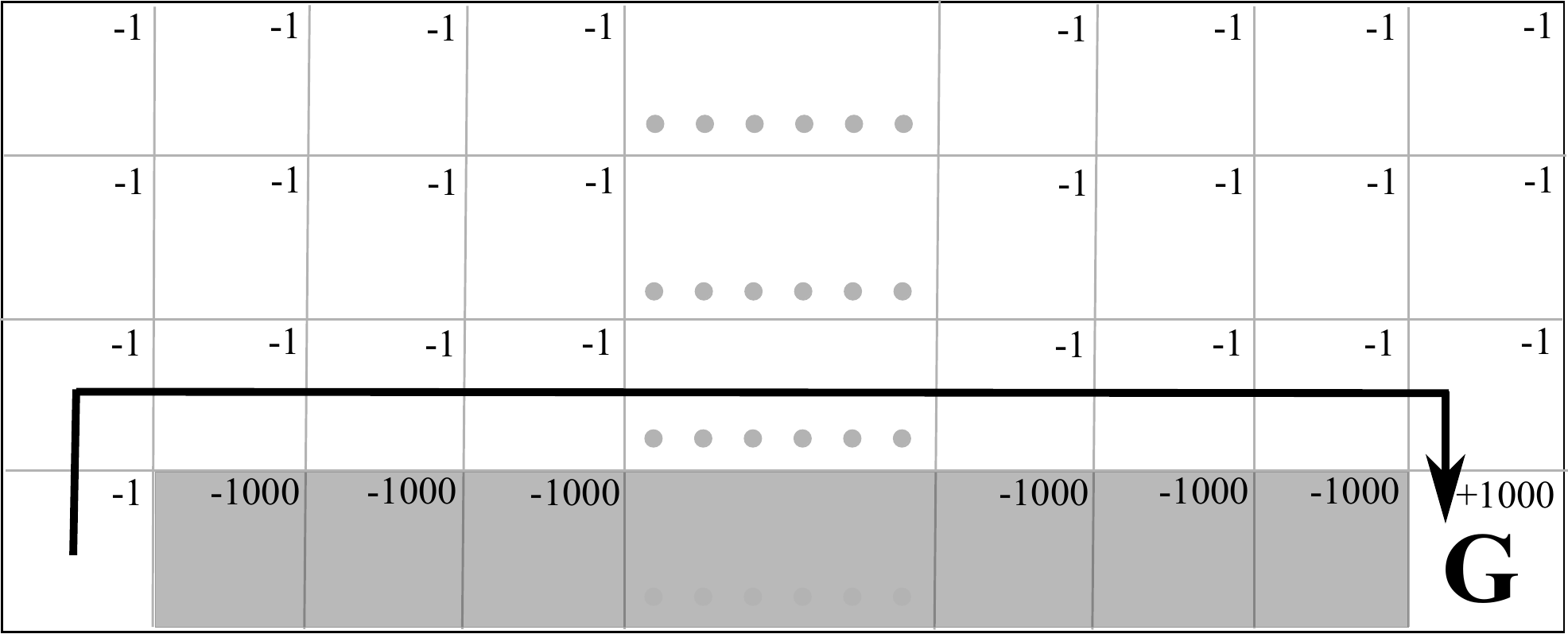}
  \label{fig:cliff:pi*_MR}}
  \subfigure[$\tau(\pi^*_{\mathcal{M}_R^H})$]
  {\includegraphics[scale=0.25]{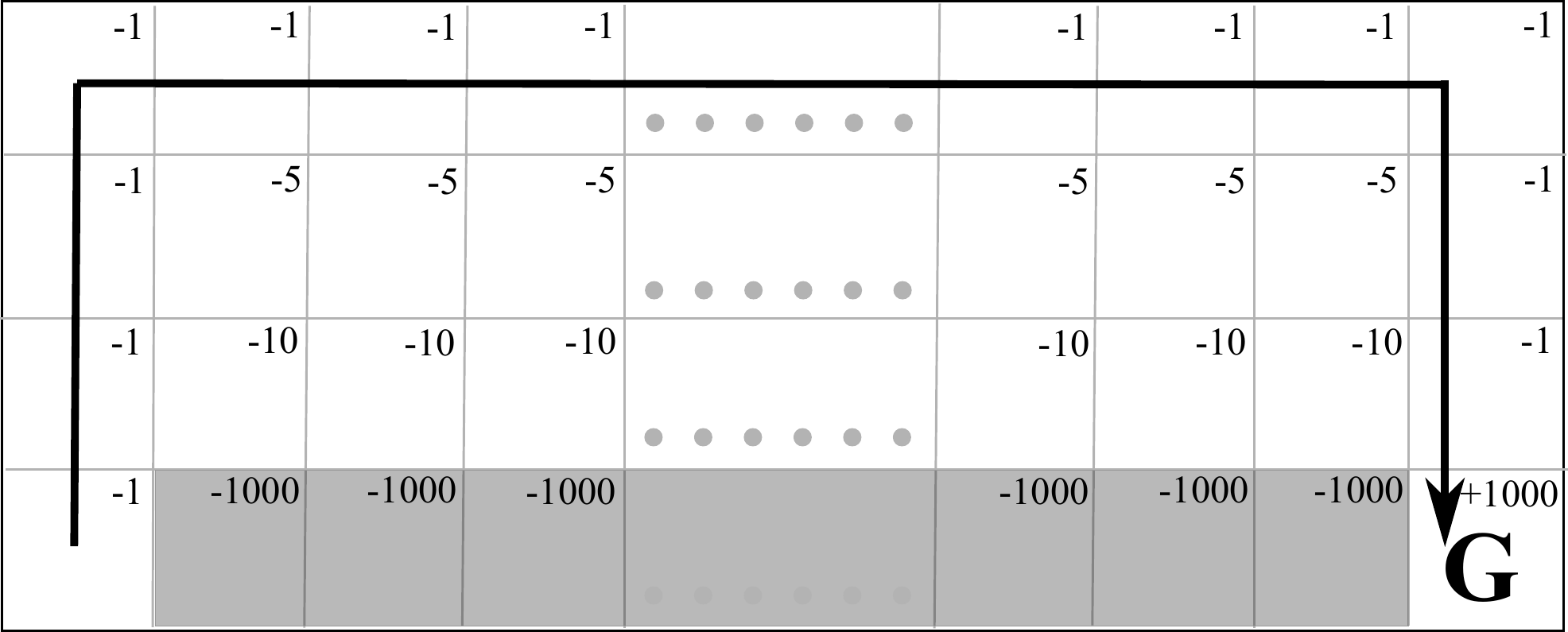}
  \label{fig:cliff:pi*_MH}}
  \subfigure[$\tau(\pi^*_\mathcal{E}) | \delta = 0.95$]
  {\includegraphics[scale=0.25]{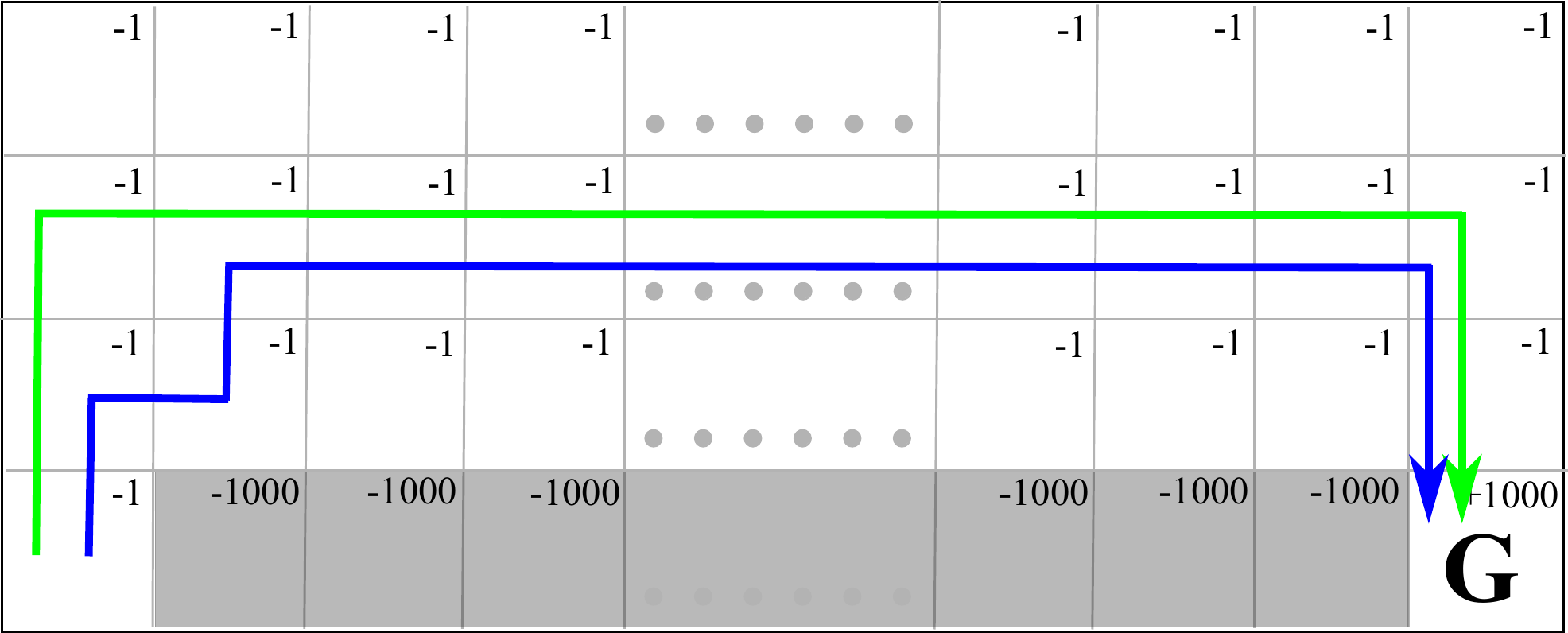}
  \label{fig:cliff:pi*_exp}}
  \caption{Behavior comparison in the large cliff world. Grey areas is the cliff and G is the goal. Reward for each state is shown at the top right corner. Displayed are the most likely trajectories from policies: (a) the optimal policy under $\mathcal{M}_R$, (b) the optimal policy under $\mathcal{M}_R^H$ (i.e., human expectation), (c) the safe explicable policies returned by PDT+ (green) and PAG+ (blue). 
  }
  \label{fig:cliff}
\end{figure*}

\begin{figure}[ht]
  \centering

  \subfigure[]
  {\includegraphics[scale=0.26]{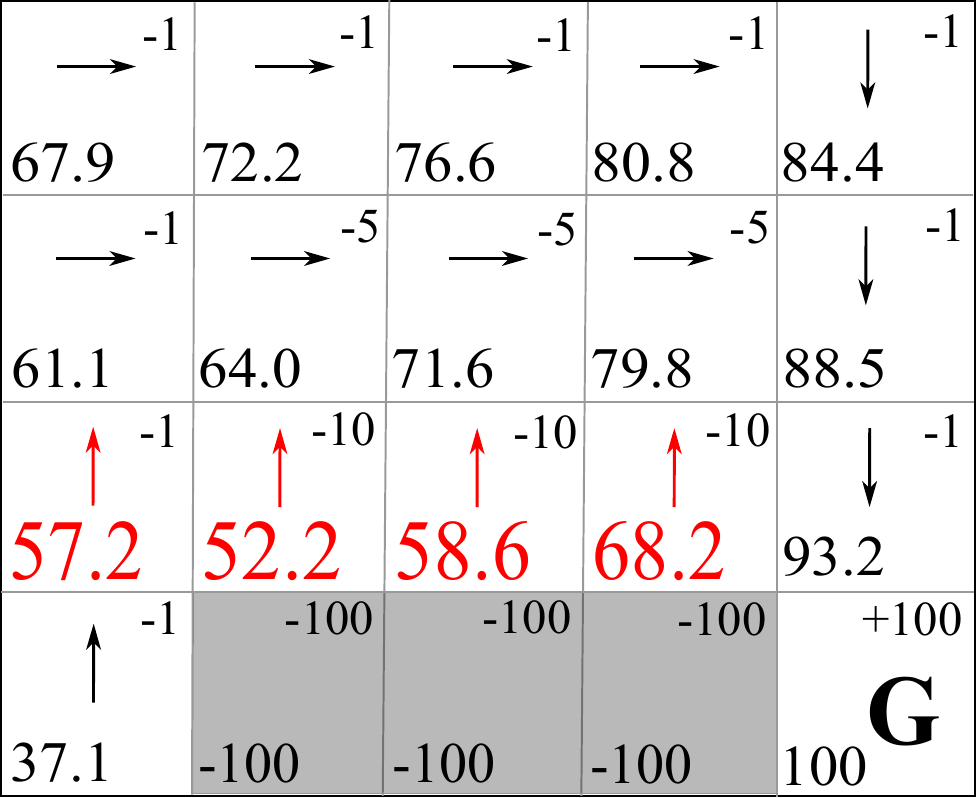}
  \label{fig:cliff-paretoSet:p1}}
  \subfigure[]
  {\includegraphics[scale=0.26]{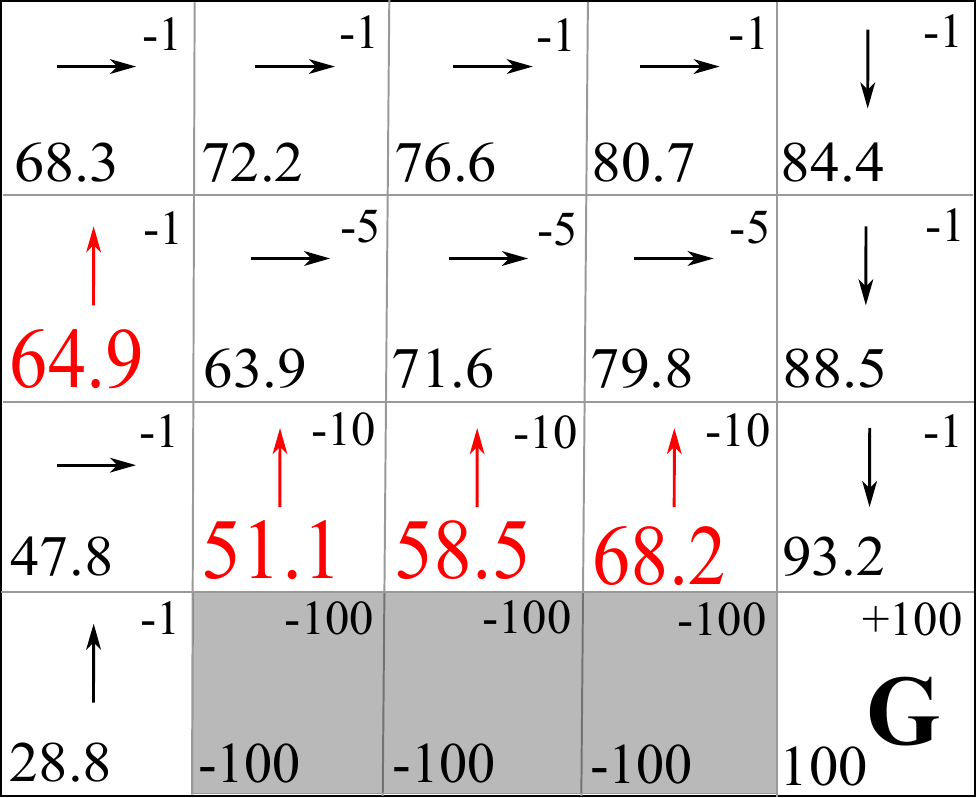}
  \label{fig:cliff-paretoSet:p2}}
  \subfigure[]
  {\includegraphics[scale=0.26]{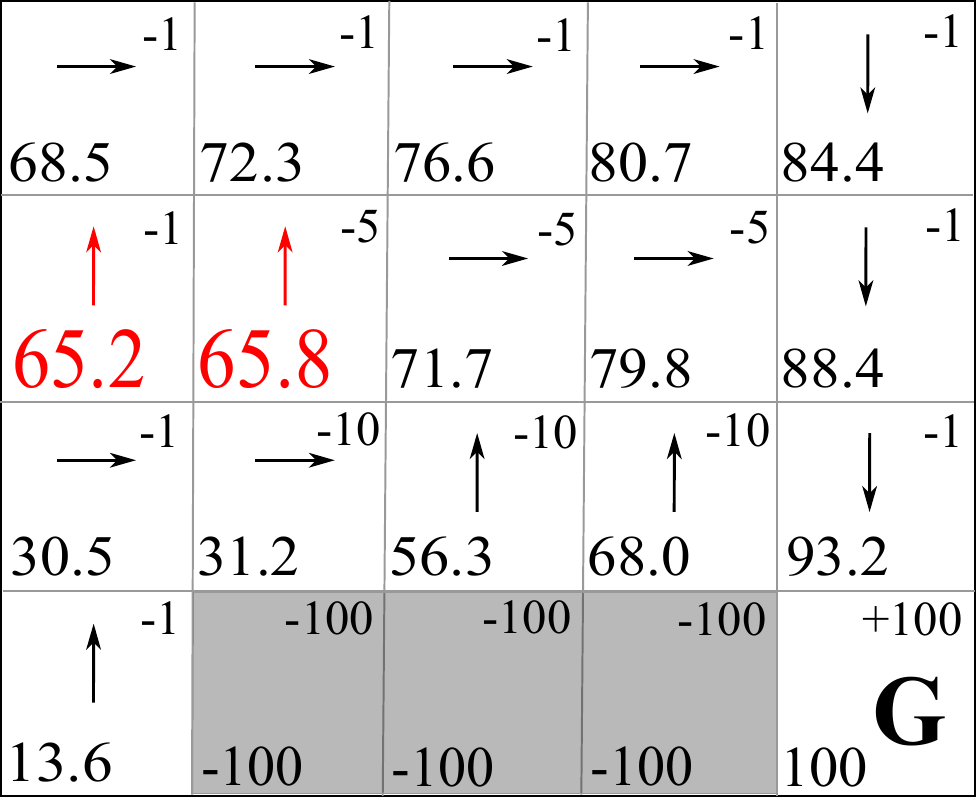}
  \label{fig:cliff-paretoSet:p3}}
  \caption{Pareto set obtained by PDT+ when $\delta=0.90$ and their corresponding $V$ values in $\mathcal{M}_R^H$, in the small cliff world. Values highlighted in red are those that result in non-dominated policies. 
  (b) shows the policy obtained by PAG+.
  }
  \label{fig:cliff-paretoSet}
\end{figure}

$2)$ {\it Behavior Comparison in Cliff Worlds}:
The results of the cliff worlds are shown in Figs. \ref{fig:cliff-paretoSet} (CS) and \ref{fig:cliff} (CL). 
Both the small and large domains introduce similar behaviors (shown only in the large domain): the optimal behavior in the agent's model takes the shortest path (Fig. \ref{fig:cliff:pi*_MR}), whereas that in the human's model stays as far away from the cliff as possible (Fig. \ref{fig:cliff:pi*_MH}). 
For SEP, Fig. \ref{fig:cliff-paretoSet} shows all the three policies in the Pareto set obtained given $\delta=0.90$ in the small domain. 
Fig. \ref{fig:cliff:pi*_exp} shows the most likely trajectories resulting from the policies in the Pareto set obtained given $\delta = 0.95$ in the large domain using the approximate solution. 
In general, we observe that the safe explicable policies result in trajectories that steer the agent away but not too far from the cliff to satisfy the bound while aligning with the human's expectation. 
In cliff worlds, to choose from $\Pi^*_{\mathcal{E}}$, 
we assign higher scores to policies producing simpler behaviors (e.g., fewer turns),
it led to choosing the policy producing the green trajectory in Fig. \ref{fig:cliff:pi*_exp} and the policy in Fig. \ref{fig:cliff-paretoSet:p1}.
PAGs, on the other hand, computed different policies in $\Pi^*_{\mathcal{E}}$ (see figures). 

$3)$ {\it Behavior Comparison in Wumpus World}:
The results are shown in Fig. \ref{fig:wumpus}. 
Following the optimal policy in the agent's model ($\mathcal{M}_R$), the agent collected both coins while staying near the wumpus before exiting, as shown in Fig. \ref{fig:wump:pi*_R}. 
Following the optimal policy in the human's model ($\mathcal{M}_R^H$), the agent avoided getting close to the wumpus and collected a single coin before exiting, as shown in Fig. \ref{fig:wump:pi*_H}.
When applying SEP under the bound $\delta = 0.90$, PDT+ returns a large Pareto set (see Tab. \ref{tab:ablation&runtime}). 
To select from $\Pi^*_{\mathcal{E}}$, we score policies based on the average distance between the agent and the wumpus throughout the most likely trajectory. 
The trajectory from the policy with the highest score is shown in Fig. \ref{fig:wump-PDT}: we can observe that the agent managed to collect both coins while maintaining a cautious distance from the wumpus, albeit taking a longer path, which is more explicable than the agent's optimal behavior in Fig. \ref{fig:wump:pi*_R} and simultaneously more efficient than the human's expectation in Fig. \ref{fig:wump:pi*_H}. Fig. \ref{fig:wump-PAG} showcases the behavior obtained by PAG+, which also maintains a cautious distance from the wumpus, for the most part.

%%%%% CLIFF EXPERIMENT PICTURES

\subsection{Physical Robot Experiment}
\subsubsection{Robot Assistant Domain:}
We implemented a scenario similar to the motivating example, where a MOVO robot assists a human user in setting up the dining table (Fig. \ref{fig:movo-exp}). 
The task involves fetching a napkin for the user from another table. 
However, the user lacks a full understanding of the kinematic constraints of the robot arms, expecting the robot to reach any location within its arm's length. 
Consequently,
the user anticipates that the robot will place the napkin beside the
plate, close to her.
% Consequently, the user anticipates the robot to place the napkin next to the plate, close to her.
In contrast, the robot's model accounts for restricted arm movement due to a vase on the table. Placing the napkin close to the user may risk tipping over the vase containing water, posing a safety hazard. Therefore, the robot's optimal behavior dictates placing the napkin next to the vase, albeit farther away from the user.

In this experiment, we operated within a discretized environment where the state space was defined by the following variables: the robot's location, the napkin's location, and the vase's location. Transitions between discrete states were facilitated by pre-generated robot trajectories using {\it Move It}.
Specifically, in $\mathcal{M}_R^H$ the robot can access any location on the dining table regardless of its own position or the vase's placement, whereas $\mathcal{M}_R$ accurately reflects the influences from these factors.
Our objective is to showcase that a robot operating under SEP would opt for a costlier policy in $\mathcal{M}_R$ to align with human expectations while ensuring safety.

%%%%%%%%%  Robot Experiment
\iffalse
\begin{figure}[ht]
  \subfigure
  {\includegraphics[scale=0.15]{Images/movo/exp1_pick.png}
  \label{fig:movo-exp1_pick}}
  \hspace*{\fill}
  \subfigure
  {\includegraphics[scale=0.15]{Images/movo/exp2_clear.png}
  \label{fig:movo-exp2_clear}}
  
  \subfigure
  {\includegraphics[scale=0.15]{Images/movo/exp1_move.png}
  \label{fig:movo-exp1_move}}
  \hspace*{\fill}
  \subfigure
  {\includegraphics[scale=0.15]{Images/movo/exp2_pick.png}
  \label{fig:movo-exp2_pick}}
  
  \subfigure
  {\includegraphics[scale=0.15]{Images/movo/exp1_place.png}
  \label{fig:movo-exp1_place}}
  \hspace*{\fill}
  \subfigure
  {\includegraphics[scale=0.15]{Images/movo/exp2_place.png}
  \label{fig:movo-exp2_place}}
  \caption{Behavior generated by $\pi^*_{\mathcal{E}}$ under $\delta=0.15$. In (a), the robot picks the napkin. In (b), as shown by the arrow, the robot moves (translation movement) from its original position to a position closer to the human in order to avoid tipping the vase over. In (c), the manipulator places the napkin close to the human. Notice the displacement of the robot's position in (b) and (c) with respect to its position in (a). 
  Behavior generated by $\pi^*_{\mathcal{E}}$ under $\delta=0.20$. Sub-figures (a), (b)  and (c) show the steps taken by the robot. In (a), the robot clears the obstacle (flower vase) from its path. In (b), the manipulator picks the napkin, and in (c), the manipulator places the napkin close to the human.} 
  \label{fig:movo-exp}
\end{figure}
\fi
%%%%%%

\subsubsection{Results:} 

\begin{figure}[ht!]
    \centering
    \subfigure[$\pi^*_{\mathcal{E}}$ $|$ $\delta=0.85$]
    {\includegraphics[scale=0.39]{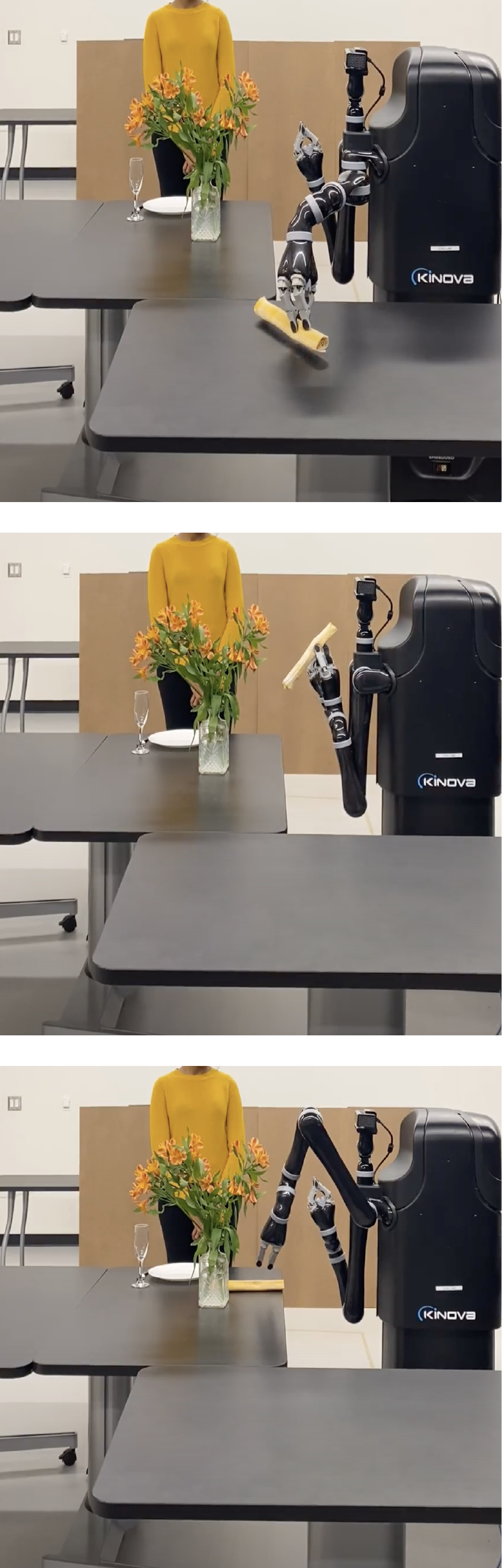}
    \label{fig:movo-exp1}}
    % \hspace*{\fill}
    \hspace{20pt}
    \subfigure[$\pi^*_{\mathcal{E}}$ $|$ $\delta=0.80$]
    {\includegraphics[scale=0.39]{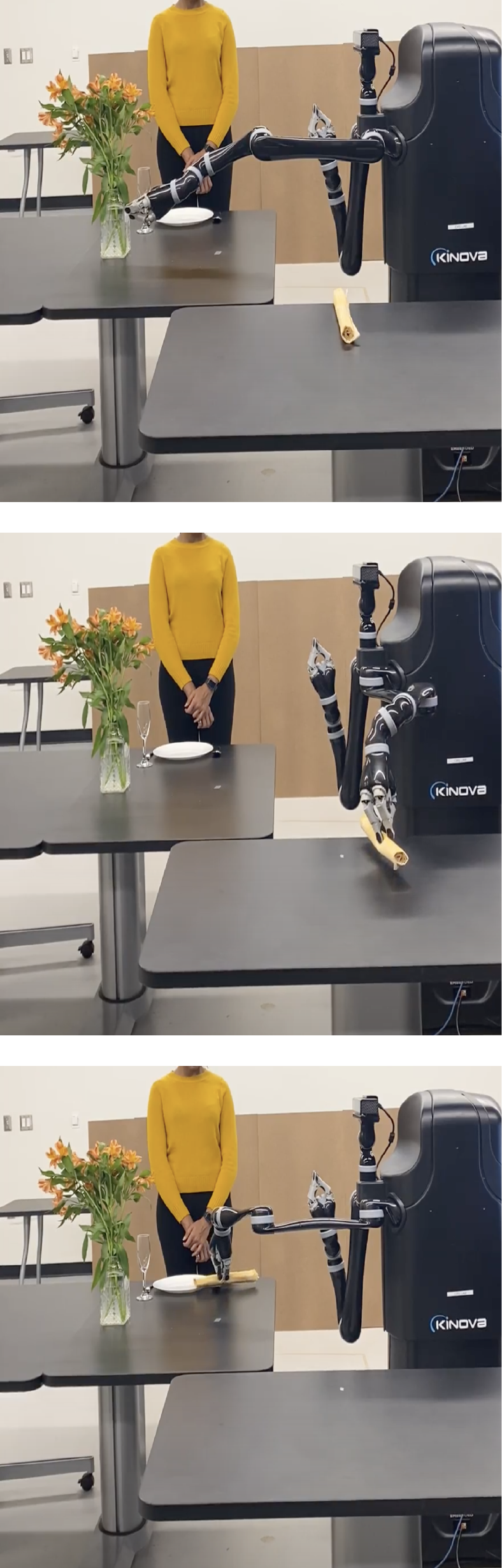}
    \label{fig:movo-exp2}}
    \caption{Safe explicable behaviors generated by PAG+ in the robot assistant domain under different bounds. 
    }
    \label{fig:movo-exp}
\end{figure}

Fig. \ref{fig:movo-exp} depicts the safe explicable behaviors obtained from the robot experiment. 
The optimal behavior in $\mathcal{M}_R$ had two steps: the robot picked the napkin and placed it on the table next to the vase, away from the user.
Further, we ran SEP with two different bounds, yielding two distinct safe explicable behaviors. 
When $\delta=0.85$ (Fig. \ref{fig:movo-exp1}), the robot picked the napkin, circumvented the obstruction by the vase by moving its entire base closer to the user and then placed the napkin next to the plate. 
When $\delta=0.80$ (Fig. \ref{fig:movo-exp2}), the robot initially moved the vase away to clear the obstruction, then picked the napkin and placed it next to the plate.

\section{CONCLUSIONS}

In this paper, we introduced the Safe Explicable Planning (SEP) problem, an extension of the explicable planning problem to support a safety bound. 
Our formulation generalizes the consideration of multiple objectives that are addressed in conventional MOMDPs or CMDPs to multiple models. 
The solution to SEP is a safe explicable policy that satisfies the safety bound while maximizing explicability.
We proposed an action pruning technique to reduce the search space, an exact method to find the Pareto set of policies, and a greedy method to find a single policy in the Pareto set. 
We discussed approximate solutions through state aggregation based on state features and action choices to address scalability.
However, our methods are still susceptible to policy explosion in complex domains our approach shows initial steps towards finding approximate safe explicable policies, with further research needed for more generalized and efficient approximation solutions. We conducted evaluations via simulations and physical robot experiments to validate the efficacy of our approach.

\section{Acknowledgments}
This research is supported in part by the NSF grant 2047186. The authors would also like to thank the anonymous reviewers for their helpful comments and suggestions.

\bibliography{aaai24}



\section*{Supplementary material}

# Appendices for Safe Explicable Planning

The document includes four appendices. First, the complete proofs for Lemma. 1, Lemma. 2, Theorem 1, and Theorem 2 are presented in Appendix A. Second, the domain descriptions for the small cliff world (CS), large cliff world (CL), wumpus world (W), and the physical robot experiments are presented in Appendix B. Appendix B also includes the link to the demo of the physical robot experiments. Third, the link to the code of SEP is present in Appendix C. Lastly, additional discussions and conclusions are presented in Appendix D.

# 1   Appendix A: Proofs

**Lemma 1.** *The set of policies after action pruning based on Eqn. (4) is a superset of the set of policies that satisfy the constraint in Eqn. (2), i.e., $\widetilde{\Pi} \supseteq \Pi_\delta$.*

*Proof.* To prove this we show that an action pruned by any state per Eqn. (4) is guaranteed to introduce policies that do not satisfy the constraint in Eqn. (2). Consider *any* state $s \in \mathcal{S}$ and *any* pruned action $a' \in \mathcal{A}(s) \backslash \widetilde{\mathcal{A}}(s)$ in that state. Consider choosing $a'$ in $s$ once and thereafter choosing actions as per the optimal policy. The expectation for this is given by

$$Q^{\pi^*}_{\mathcal{M}_R}(s, a') = \mathbb{E}^{\pi^*}_{\mathcal{T}_R}[r_R(s^1) + \gamma_R V^{\pi^*}_{\mathcal{M}_R}(s^1) | s^0 = s, a^0 = a'].$$

From Eqn. (4), w.k.t.

$$Q^{\pi^*}_{\mathcal{M}_R}(s, a') < \delta \mathbb{E}^{\pi^*}_{\mathcal{T}_R}[r_R(s^1) + \gamma_R V^{\pi^*}_{\mathcal{M}_R}(s^1) | s^0 = s, a^0 = \pi^*(s)]$$

Since the future states already choose the optimal actions, there is no room to improve the value of $Q^{\pi^*}_{\mathcal{M}_R}(s, a')$ and hence cannot satisfy $\delta$. Thus, any policy that chooses $a'$ for $s$ cannot satisfy the constraint in that state. □

**Lemma 2.** *Let $\pi$ and $\pi'$ be two deterministic policies that differ by only a single action in some state i.e., $\exists s_i \in \mathcal{S}\ [\pi'(s_i) \neq \pi(s_i)] \wedge \forall s_j \in \mathcal{S} \setminus \{s_i\}\ [\pi'(s_j) = \pi(s_j)]$ and satisfy $Q^{\pi}_{\mathcal{M}_R}(s_i, \pi'(s_i)) \leq V^{\pi}_{\mathcal{M}_R}(s_i)$. Then, policy $\pi'$ is a descendant of $\pi$ in PDT, i.e., policy $\pi'$ is no better than $\pi$, or more formally, $\forall s \in \mathcal{S}\ [V^{\pi'}_{\mathcal{M}_R}(s) \leq V^{\pi}_{\mathcal{M}_R}(s)]$.*

*Proof.* This is an extension of the policy improvement theorem (Sutton and Barto 2018) but in the opposite direction (hence referred to as a policy descent). We know that $\forall s \in \mathcal{S}$,

$$Q^{\pi}_{\mathcal{M}_R}(s, \pi(s)) = V^{\pi}_{\mathcal{M}_R}(s)$$
$$\mathbb{E}^{\pi}_{\mathcal{T}_R}[r_R(s^1) + \gamma_R V^{\pi}_{\mathcal{M}_R}(s^1) | s^0 = s, a^0 = \pi(s)] = V^{\pi}_{\mathcal{M}_R}(s)$$

1

Consider a temporary non-stationary policy $\pi_1'$ that chooses the action as per $\pi'$ once in the initial state $s^0$ and follows $\pi$ thereafter. $\forall s \in \mathcal{S}$,

$$Q_{\mathcal{M}_R}^{\pi_1'}(s, \pi_1'(s)) \leq Q_{\mathcal{M}_R}^{\pi}(s, \pi(s))$$
$$Q_{\mathcal{M}_R}^{\pi_1'}(s, \pi_1'(s)) \leq V_{\mathcal{M}_R}^{\pi}(s)$$
$$\mathbb{E}_{\mathcal{T}_R}^{\pi_1'}[r_R(s^1) + \gamma_R V_{\mathcal{M}_R}^{\pi_1'}(s^1) | s^0 = s, a^0 = \pi_1'(s)] \leq V_{\mathcal{M}_R}^{\pi}(s)$$

This is because $s^0$ can either be $s_i \in \mathcal{S}$ or any $s_j \in \mathcal{S} \setminus \{s_i\}$.
If $s^0 = s_i$ then, $Q_{\mathcal{M}_R}^{\pi}(s_i, \pi'(s_i)) \leq V_{\mathcal{M}_R}^{\pi}(s_i)$ (given).
If $s^0 = s_j$ then, $Q_{\mathcal{M}_R}^{\pi}(s_j, \pi'(s_j)) = V_{\mathcal{M}_R}^{\pi}(s_j)$ as $\pi_1'$ differs from $\pi$ only in one action in one state $s_i$ and follows $\pi$ in all future states.

Similarly, consider another temporary non-stationary policy $\pi_2'$ that chooses the actions as per $\pi'$ once in $s^0$, again in $s^1$, and follows $\pi$ thereafter. $\forall s \in \mathcal{S}$,

$$Q_{\mathcal{M}_R}^{\pi_2'}(s, \pi_2'(s)) \leq Q_{\mathcal{M}_R}^{\pi_1'}(s, \pi_1'(s))$$
$$Q_{\mathcal{M}_R}^{\pi_2'}(s, \pi_2'(s)) \leq Q_{\mathcal{M}_R}^{\pi}(s, \pi(s))$$
$$\mathbb{E}_{\mathcal{T}_R}^{\pi_2'}[r_R(s^1) + \gamma_R V_{\mathcal{M}_R}^{\pi_2'}(s^1) | s^0 = s, a^0 = \pi_2'(s)] \leq V_{\mathcal{M}_R}^{\pi}(s)$$

Consider repeating this until we always choose actions as per $\pi'$.

$$\mathbb{E}_{\mathcal{T}_R}^{\pi'}[r_R(s^1) + \gamma_R r_R(s^2) + \gamma_R^2 r_R(s^3) + ... | s^0 = s, a^0 = \pi'(s)] \leq V_{\mathcal{M}_R}^{\pi}(s)$$
$$V_{\mathcal{M}_R}^{\pi'}(s) \leq V_{\mathcal{M}_R}^{\pi}(s)$$

□

**Theorem 1.** *PDT+ returns all Pareto optimal policies in $\Pi_{\mathcal{E}}^*$.*

*Proof.* To prove this, we show that there exists a policy descent path from any optimal policy (denoted by $\pi^*$) in $\mathcal{M}_R$ (i.e., the root node in PDT) to any Pareto optimal policy (denoted by $\pi_{\mathcal{E}}^*$) in $\Pi_{\mathcal{E}}^*$ by induction. Let $n$ denote the number of actions a policy differs from $\pi^*$.

When $n = 1$, i.e. $\pi_{\mathcal{E}}^*$ differs from $\pi^*$ in a single action i.e. $\exists s_i \in \mathcal{S} \ [\pi_{\mathcal{E}}^*(s_i) \neq \pi^*(s_i)] \land \forall s_j \in \mathcal{S} \setminus \{s_i\} \ [\pi_{\mathcal{E}}^*(s_i) = \pi^*(s_i)]$ then, $\forall s \in \mathcal{S} \ Q_{\mathcal{M}_R}^{\pi^*}(s, \pi_{\mathcal{E}}^*(s)) \leq V_{\mathcal{M}_R}^{\pi^*}(s)$ as $\pi^*$ is the optimal policy. This makes $\pi_{\mathcal{E}}^*$ one of the direct descendants of $\pi^*$ (by Lem. 2). Hence, $\pi_{\mathcal{E}}^*$ is expanded by PDT.

When $n = k$, i.e. $\pi_{\mathcal{E}}^*$ differs from $\pi^*$ in any $k$ actions i.e. $\exists \mathcal{S}_k \subseteq \mathcal{S} \ \forall s_i \in \mathcal{S}_k \ [\pi_{\mathcal{E}}^*(s_i) \neq \pi^*(s_i)] \land \forall s_j \in \mathcal{S} \setminus \mathcal{S}_k \ [\pi_{\mathcal{E}}^*(s_j) = \pi^*(s_j)]$, assume $\pi_{\mathcal{E}}^*$ is expanded by PDT.

When $n = k+1$, i.e. $\pi_{\mathcal{E}}^*$ differs from $\pi^*$ in any $k+1$ actions i.e. $\exists \mathcal{S}_{k+1} \subseteq \mathcal{S} \ \forall s_i \in \mathcal{S}_{k+1} \ [\pi_{\mathcal{E}}^*(s_i) \neq \pi^*(s_i)] \land \forall s_j \in \mathcal{S} \setminus \mathcal{S}_{k+1} \ [\pi_{\mathcal{E}}^*(s_j) = \pi^*(s_j)]$, there must be at least one action out of the $k+1$ actions aligning with $\pi^*$ that improves, or is same as, the value of $\pi_{\mathcal{E}}^*$.
Assume this is false i.e. all the $k+1$ actions aligning with $\pi^*$ worsen $\pi_{\mathcal{E}}^*$. The policy introduced by aligning all $k+1$ actions with $\pi^*$ is $\pi^*$ itself (as all actions other than the $k+1$ actions in $\pi_{\mathcal{E}}^*$ are same as $\pi^*$). W.k.t. $\pi^*$ is optimal and cannot be worse than $\pi_{\mathcal{E}}^*$, which is a contradiction. Thus, $\exists \pi \ \exists s_i \in \mathcal{S}_{k+1} \ [\pi_{\mathcal{E}}^*(s_i) \neq \pi(s_i) = \pi^*(s_i)] \land \forall s_j \in \mathcal{S} \setminus \{s_i\} \ [\pi(s_j) = \pi_{\mathcal{E}}^*(s_j)]$ that satisfies $Q_{\mathcal{M}_R}^{\pi_{\mathcal{E}}^*}(s_i, \pi(s_i)) \geq V_{\mathcal{M}_R}^{\pi_{\mathcal{E}}^*}(s_i)$ and by Lem. 2, $\pi$ is no-worse than $\pi_{\mathcal{E}}^*$. W.k.t. $\pi$ is

expanded (induction assumption). Consequently, $\pi_\mathcal{E}^*$ must be one of the direct descendants of $\pi$ in PDT and hence $\pi_\mathcal{E}^*$ is expanded.

Therefore, the result holds for any $n$ by the principle of induction. Finally, the same conclusion holds for PDT+ (by Lem. 1). □

**Theorem 2.** *PAG+ returns a policy in the Pareto set $\Pi_\mathcal{E}^*$.*

*Proof.* The PAG search process stops when it can no longer improve or find a policy that is equivalent in values to $\pi_\mathcal{E}$ under $\mathcal{M}_R^H$ while satisfying the safety constraint. This translates to that there does not exist a state-action update that implements a policy ascent step under the constraint i.e., $\neg\exists(s' \in \mathcal{S}, a' \in \mathcal{A})$
$[Q_{\mathcal{M}_R^H}^{\pi_\mathcal{E}}(s', a'=\pi'(s)) \geq Q_{\mathcal{M}_R^H}^{\pi_\mathcal{E}}(s', \pi_\mathcal{E}(s'))$, s.t. $\forall s' \in \mathcal{S}\ [V_{\mathcal{M}_R}^{\pi'}(s') \geq \delta V_{\mathcal{M}_R}^{\pi^*}(s')]]$. However, if $\pi_\mathcal{E} \notin \Pi_\mathcal{E}^*$, there must exist another policy $\pi \in \Pi_\mathcal{E}^*$ that dominates $\pi_\mathcal{E}$ i.e. $\exists(s \in \mathcal{S}, a \in \mathcal{A})$
$[Q_{\mathcal{M}_R^H}^{\pi_\mathcal{E}}(s, a=\pi(s)) > Q_{\mathcal{M}_R^H}^{\pi_\mathcal{E}}(s, \pi_\mathcal{E}(s))$, s.t. $\forall s \in \mathcal{S}\ [V_{\mathcal{M}_R}^{\pi}(s) \geq \delta V_{\mathcal{M}_R}^{\pi^*}(s)]]$. This contradicts with the fact that no policy ascent step exists.
Therefore, $\pi_\mathcal{E} \in \Pi_\mathcal{E}^*$.
Finally, the same conclusion holds for PAG+ (by Lem. 1).

□

## 2 Appendix B: Domain Descriptions

### 2.1 Simulations

1) Cliff Worlds (CS & CL): In the cliff worlds, the agent is required to navigate alongside the edge of a cliff to reach the goal. We created a small $4 \times 5$ grid, referred to as CS (shown in Fig. 5) to evaluate exact methods and a large $4 \times 100$ grid, referred to as CL (shown in Fig. 4) to evaluate approximate methods. To apply approximate methods to CL, the states were aggregated based on features such as distance to the cliff, and agent's position in the grid (e.g., along the edge or at the ends). For CL, we aggregated all non-terminal states into 10 clusters and retained the terminal states (cliffs and goal) as is.

The ground truth ($\mathcal{M}_R$) is that the agent can traverse alongside the edge without slipping off the cliff. Accordingly, $\mathcal{T}_R$ is designed such that the agent heads in the right direction with a probability of 0.9 and remains in the same state with a probability of 0.1. The human's belief ($\mathcal{M}_R^H$) is that the agent may slip off from the edge with some probability, and the terrain closer to the cliff is more uneven and hence more difficult for the agent to traverse. Accordingly, $\mathcal{T}_R^H$ is designed such that the agent heads in the right direction with a probability of 0.7, steers in either direction with a probability of 0.1 each, and remains in the same location with a probability of 0.1.

The reward functions $\mathcal{R}_R$ and $\mathcal{R}_R^H$ are shown in Figs. 4(a) and 4(b) respectively for CL. For CS, the rewards for non-terminal states remain the same but for the cliff states it is $-100$ instead of $-1000$, and for the goal state it is 100 instead of 1000.

2) Wumpus World (W): In the wumpus world, we created a $5 \times 5$ grid referred to as W (shown in Fig. 3). To apply approximate methods to W, the non-terminal states were aggregated into 15 clusters based on features such as the relative direction of the wumpus from agent and collection status of the gold coins. The agent is required to exit the $5 \times 5$

cave while collecting gold coins on its way out and avoiding encounters with the wumpus (i.e., staying in the same location). The cave has a moving wumpus, two gold coins, and an exit location.

The ground truth ($\mathcal{M}_R$) is that the agent's movement is deterministic and the wumpus's movement is stochastic. Accordingly, $\mathcal{T}_R$ is designed such that the wumpus always chooses to move toward the agent's current location with uniform probability. The human's belief $\mathcal{M}_R^H$ is that the actions of both agents are stochastic. Under such a belief, the human would consider it dangerous for the agent to move close to the wumpus. Accordingly, $\mathcal{T}_R^H$ is designed such that the dynamics of the agent's movement are the same as that in the cliff world and the dynamics of the wumpus's movement is to move close to the agent's current location with uniform probability.

In reward functions $\mathcal{R}_R$ and $\mathcal{R}_R^H$, collecting each gold coin gives a reward of $+30$. The game terminates if the agent encounters the wumpus with a reward of $-100$ or if it exits the cave with a reward of $+100$. The living reward is $-0.1$.

## 2.2 Physical Robot Experiment

Robot Assistant Domain: In this domain, a Kinova MOVO robot is assisting a human user with setting up the dining table (refer to Fig. 6). The state was modeled to include the location of an object (which was a napkin in this experiment), the location of an obstacle (which was a vase in this experiment), and the location of the robot. The possible locations for the napkin are viz., on the side table (shown in the top sub-figures), on the front table next to the vase which is away from the human, on the front table next to the plate which is near the human (shown in the bottom sub-figures), and in the robot's gripper (shown in the middle sub-figures). The possible locations for the vase are near the robot (shown in left sub-figures) and away from the robot (shown in right sub-figures), and fallen (if tipped over which is not shown). The possible locations for the robot are close to the side table (shown in the right sub-figures) and close to the human (shown in the middle-left and middle-right sub-figures).

The robot is required to fetch a napkin for the user from another table. In the robot's model ($\mathcal{M}_R$), movement of the arms is restricted by a vase on the table such that placing the napkin close to the user may tip over the vase containing water, resulting in a safety risk. Hence, the robot's optimal behavior is to pick the napkin from the side table and place it next to the vase, which is further away from the user as shown in Fig. 7. Accordingly, $\mathcal{T}_R$ is designed such that moving the napkin from the side table directly toward the human with the obstacle in the way (w.r.t. the robot position) will result in tipping the vase with probability 1.0. However, clearing the vase could successfully displace it with a probability of 0.9, tip it over with a probability of 0.05 and leave it as is with a probability of 0.05.

The user does not fully understand the kinematic constraints of the robot arms and hence expects the robot to

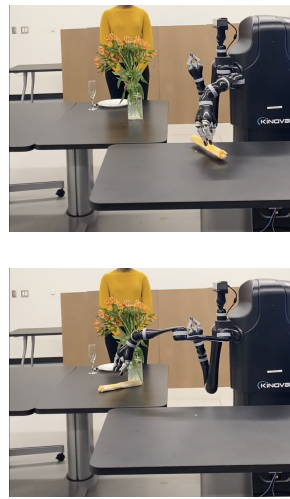

Figure 7: $\pi^*$ in $\mathcal{M}_R$

4

place the napkin next to the plate which is close to her.
Accordingly, $\mathcal{T}_R^H$ is designed such that the robot can access
any location on the table irrespective of its own location or the location of the obstacle. In both $\mathcal{T}_R$ and $\mathcal{T}_R^H$, the probability of executing an action successfully is 0.9 and the probability of failing is 0.1.

The environment terminates when the napkin is placed in any location on the table or if the flower vase is tipped over. In $\mathcal{R}_R$, there is an equal reward of 10 for placing the napkin anywhere on the table and $-10$ for hitting the vase. In $\mathcal{R}_R^H$, there is a reward of 10 for placing the napkin close to the human and 0 reward for placing the napkin anywhere else on the table.

Please refer to this demo video for further illustration.

# 3 Appendix C: Code

The code for SEP is available in this repository.

# 4 Appendix D: Discussion and Conclusions

In this paper, we introduced the problem of Safe Explicable Planning (SEP), which significantly extends explicable planning to support a safety bound. To focus on the planning challenges of introducing safety in explicable planning, we assume the human belief model $\mathcal{M}_R^H$ is known. We provide references to existing literature where a similar assumption is made. When $\mathcal{M}_R^H$ is unknown, it can be learned by querying human subjects, to learn a reward function such as the approaches discussed in the survey paper (Wirth et al. 2017) and to learn domain dynamics such as the approach by (Zhuo 2015). We defer the consideration of other forms of $\mathcal{M}_R^H$ (such as hierarchical models) or learning $\mathcal{M}_R^H$ for future work. It is worth noting that the problem of SEP can be formulated as a CMDP problem under the special case where $\mathcal{T}_R = \mathcal{T}_R^H$ and $\gamma_R = \gamma_R^H$, in planning (Altman 2021) or learning (Achiam et al. 2017) setting.

In our work, we assume that safety is correlated to the expected return in the agent's model under the intuition that unsafe behaviors would result in low returns. Thus, imposing a lower bound on the return prevents unsafe behaviors. Such a safety definition is based on the constrained criterion. We aim to extend it to consider other safety formulations such as the worst-case criterion etc. outlined in (Garcia and Fernandez 2015) in the future. It would also be interesting to study how our value function-based criterion can be compared with or potentially equated to a state-machine-based criterion such as that studied by (Hunt et al. 2021). For example, prior work has studied how LTL constraints can be approximately considered by shaping the reward function (Camacho et al. 2019; Li, Vasile, and Belta 2017). Further, the safety bound in our work is domain-specific as the criticality of safety is different in different scenarios and can be specified by the designer. We discuss how the safety bound was determined for the experimental evaluations and will more systematically address it in future work.

Our problem formulation generalizes the consideration of objectives from multiple models

and the solution is a Pareto set of policies. We proposed an action pruning technique to reduce the search space, an exact method that returns the Pareto set, and a greedy method that returns a single policy. Existing literature in MOMDP shows that finding exact Pareto solutions for complex problems is intractable. To scale to complex domains, we further discussed approximate solutions by clustering states that are alike in terms of action selection. Since SEP requires searching the policy space, the methods proposed (exact and approximate) are still susceptible to policy explosion in large domains. The aggregation-based approximation proposed is preliminary work toward finding approximate safe explicable policies. We defer the study of generalized and more efficient approximation techniques to solve SEP in the future.



\end{document}